\documentclass[11pt]{article}
\usepackage[preprint]{acl}
\usepackage{times}
\usepackage{latexsym}
\usepackage[T1]{fontenc}
\usepackage[utf8]{inputenc}
\usepackage{amsmath}
\usepackage{amssymb}
\usepackage{amsthm}
\usepackage{booktabs}
\usepackage{array}
\usepackage{multirow}
\usepackage{graphicx}
\usepackage{xspace}
\usepackage{microtype}
\usepackage{tikz}
\usetikzlibrary{arrows.meta,positioning,calc}
\definecolor{oiBlue}{HTML}{0072B2}
\definecolor{oiVerm}{HTML}{D55E00}
\definecolor{oiGreen}{HTML}{009E73}

\newcommand{\oracleDocCloudDenies}{4,504}

\newcommand{\oracleDocCloudMultiGoal}{1,900}
\newcommand{\oracleDocCloudGapFree}{0.88}
\newcommand{\oracleDocCloudGapFixed}{0.96}
\newcommand{\oracleDocCloudGapActionable}{0.998}
\newcommand{\oracleDocCloudNoRecourse}{0.15}

\newcommand{\oracleDocCloudImmutable}{\feat{is\_owner}}
\newcommand{\oracleGithubDenies}{336}

\newcommand{\oracleGithubMultiGoal}{16}
\newcommand{\oracleGithubGapFree}{0.00}
\newcommand{\oracleGithubGapFixed}{0.00}
\newcommand{\oracleGithubGapActionable}{0.00}
\newcommand{\oracleGithubNoRecourse}{0.00}
\newcommand{\oracleGithubKeptMultiGoal}{13}
\newcommand{\oracleGithubImmutable}{\feat{is\_reporter}}
\newcommand{\oracleReposDenies}{448}

\newcommand{\oracleReposMultiGoal}{192}
\newcommand{\oracleReposGapFree}{1.00}
\newcommand{\oracleReposGapFixed}{1.00}
\newcommand{\oracleReposGapActionable}{1.00}
\newcommand{\oracleReposNoRecourse}{0.00}

\newcommand{\oraclePrivilegedDenies}{276}

\newcommand{\oraclePrivilegedMultiGoal}{56}
\newcommand{\oraclePrivilegedGapFree}{0.00}
\newcommand{\oraclePrivilegedGapFixed}{1.00}
\newcommand{\oraclePrivilegedGapActionable}{1.00}
\newcommand{\oraclePrivilegedNoRecourse}{0.00}

\newcommand{\oracleSnapDenies}{3,108}

\newcommand{\oracleSnapMultiGoal}{2,176}
\newcommand{\oracleSnapGapFree}{0.96}
\newcommand{\oracleSnapGapFixed}{0.96}
\newcommand{\oracleSnapGapActionable}{0.92}
\newcommand{\oracleSnapNoRecourse}{0.45}

\newcommand{\oracleSnapImmutable}{\feat{elderly\_or\_disabled\_member} and \feat{citizenship}}
\newcommand{\oracleTitleFourDenies}{3,428}

\newcommand{\oracleTitleFourMultiGoal}{3,246}
\newcommand{\oracleTitleFourGapFree}{0.99}
\newcommand{\oracleTitleFourGapFixed}{0.999}
\newcommand{\oracleTitleFourGapActionable}{1.00}
\newcommand{\oracleTitleFourNoRecourse}{0.70}

\newcommand{\oracleTitleFourImmutable}{\feat{has\_baccalaureate\_degree} and \feat{citizenship}}
\newcommand{\oracleHudDenies}{2,292}

\newcommand{\oracleHudMultiGoal}{2,204}
\newcommand{\oracleHudGapFree}{1.00}
\newcommand{\oracleHudGapFixed}{1.00}
\newcommand{\oracleHudGapActionable}{1.00}
\newcommand{\oracleHudNoRecourse}{0.86}

\newcommand{\oracleHudImmutable}{\feat{methamphetamine\_production\_conviction}, \feat{lifetime\_sex\_offender\_registrant} and \feat{immigration\_status}}
\newcommand{\oracleNPorts}{seven}
\newcommand{\oracleNGapPorts}{six}
\newcommand{\oracleGapMin}{0.96}
\newcommand{\oracleGapMax}{1.00}
\newcommand{\oracleSynGapFree}{0.70}
\newcommand{\oracleNoRecourseEligMin}{0.45}
\newcommand{\oracleNoRecourseEligMax}{0.86}

\newcommand{\synLlamaEightFirstValid}{0.79}

\newcommand{\synLlamaEightFirstTwoPlus}{0.14}

\newcommand{\synLlamaEightFirstRecourse}{0.49}

\newcommand{\synLlamaEightFirstFidelity}{0.46}

\newcommand{\synLlamaEightNeutralTwoPlus}{0.20}
\newcommand{\synLlamaEightNeutralTwoPlusPct}{20\%}
\newcommand{\synLlamaEightNeutralRecourse}{0.45}

\newcommand{\synLlamaEightNeutralConc}{0.52}
\newcommand{\synLlamaEightNeutralConcNull}{0.29}
\newcommand{\synLlamaEightNeutralConcP}{$<$0.001}

\newcommand{\synLlamaEightAllTwoPlus}{0.75}
\newcommand{\synLlamaEightAllTwoPlusPct}{75\%}
\newcommand{\synLlamaEightAllRecourse}{0.90}

\newcommand{\synLlamaEightFirstRecourseOrderZero}{0.32}
\newcommand{\synLlamaEightFirstRecourseOrderOne}{0.65}
\newcommand{\synLlamaEightFirstListed}{0.54}
\newcommand{\synLlamaEightFirstListedChance}{0.36}
\newcommand{\synLlamaEightSwitch}{0.47}
\newcommand{\synGemmaTwelveFirstValid}{0.75}

\newcommand{\synGemmaTwelveFirstTwoPlus}{0.21}

\newcommand{\synGemmaTwelveFirstRecourse}{0.54}

\newcommand{\synGemmaTwelveFirstFidelity}{0.36}

\newcommand{\synGemmaTwelveNeutralTwoPlus}{0.34}
\newcommand{\synGemmaTwelveNeutralTwoPlusPct}{34\%}
\newcommand{\synGemmaTwelveNeutralRecourse}{0.66}

\newcommand{\synGemmaTwelveNeutralConc}{0.44}
\newcommand{\synGemmaTwelveNeutralConcNull}{0.23}
\newcommand{\synGemmaTwelveNeutralConcP}{$<$0.001}

\newcommand{\synGemmaTwelveAllTwoPlus}{0.71}
\newcommand{\synGemmaTwelveAllTwoPlusPct}{71\%}
\newcommand{\synGemmaTwelveAllRecourse}{0.99}

\newcommand{\synGemmaTwelveFirstRecourseOrderZero}{0.48}
\newcommand{\synGemmaTwelveFirstRecourseOrderOne}{0.61}
\newcommand{\synGemmaTwelveFirstListed}{0.39}
\newcommand{\synGemmaTwelveFirstListedChance}{0.37}
\newcommand{\synGemmaTwelveSwitch}{0.28}
\newcommand{\synQwenThreeFirstValid}{0.91}

\newcommand{\synQwenThreeFirstTwoPlus}{0.28}

\newcommand{\synQwenThreeFirstRecourse}{0.50}

\newcommand{\synQwenThreeFirstFidelity}{0.56}

\newcommand{\synQwenThreeNeutralTwoPlus}{0.80}

\newcommand{\synQwenThreeNeutralRecourse}{0.89}

\newcommand{\synQwenThreeNeutralConc}{0.30}
\newcommand{\synQwenThreeNeutralConcNull}{0.11}
\newcommand{\synQwenThreeNeutralConcP}{$<$0.001}

\newcommand{\synQwenThreeAllTwoPlus}{0.91}

\newcommand{\synQwenThreeAllRecourse}{0.995}

\newcommand{\synQwenThreeFirstRecourseOrderZero}{0.45}
\newcommand{\synQwenThreeFirstRecourseOrderOne}{0.67}
\newcommand{\synQwenThreeFirstListed}{0.55}
\newcommand{\synQwenThreeFirstListedChance}{0.38}
\newcommand{\synQwenThreeSwitch}{0.30}
\newcommand{\synPhiFourFirstValid}{0.97}

\newcommand{\synPhiFourFirstTwoPlus}{0.50}

\newcommand{\synPhiFourFirstRecourse}{0.67}

\newcommand{\synPhiFourFirstFidelity}{0.59}

\newcommand{\synPhiFourNeutralTwoPlus}{0.91}

\newcommand{\synPhiFourNeutralRecourse}{0.94}

\newcommand{\synPhiFourNeutralConc}{0.08}
\newcommand{\synPhiFourNeutralConcNull}{0.08}
\newcommand{\synPhiFourNeutralConcP}{0.77}

\newcommand{\synPhiFourAllTwoPlus}{0.97}

\newcommand{\synPhiFourAllRecourse}{0.999}

\newcommand{\synPhiFourFirstListed}{0.36}
\newcommand{\synPhiFourFirstListedChance}{0.40}
\newcommand{\synPhiFourSwitch}{0.09}
\newcommand{\synQwenTwoFiveFirstValid}{0.99}

\newcommand{\synQwenTwoFiveFirstTwoPlus}{0.69}

\newcommand{\synQwenTwoFiveFirstRecourse}{0.78}

\newcommand{\synQwenTwoFiveFirstFidelity}{0.52}

\newcommand{\synQwenTwoFiveNeutralTwoPlus}{0.91}

\newcommand{\synQwenTwoFiveNeutralRecourse}{0.88}

\newcommand{\synQwenTwoFiveNeutralConc}{0.20}
\newcommand{\synQwenTwoFiveNeutralConcNull}{0.10}
\newcommand{\synQwenTwoFiveNeutralConcP}{$<$0.001}

\newcommand{\synQwenTwoFiveAllTwoPlus}{0.93}

\newcommand{\synQwenTwoFiveAllRecourse}{0.97}

\newcommand{\synQwenTwoFiveFirstListed}{0.41}
\newcommand{\synQwenTwoFiveFirstListedChance}{0.44}
\newcommand{\synQwenTwoFiveSwitch}{0.16}
\newcommand{\synHaikuFirstValid}{0.99}

\newcommand{\synHaikuFirstTwoPlus}{0.50}

\newcommand{\synHaikuFirstRecourse}{0.75}

\newcommand{\synHaikuFirstFidelity}{0.95}
\newcommand{\synHaikuFirstFidelityPct}{95\%}

\newcommand{\synHaikuNeutralTwoPlus}{0.98}

\newcommand{\synHaikuNeutralRecourse}{1.00}

\newcommand{\synHaikuNeutralConc}{0.01}
\newcommand{\synHaikuNeutralConcNull}{0.07}
\newcommand{\synHaikuNeutralConcP}{$<$0.001}

\newcommand{\synHaikuAllTwoPlus}{0.99}

\newcommand{\synHaikuAllRecourse}{1.00}

\newcommand{\synHaikuFirstRecourseOrderZero}{0.71}
\newcommand{\synHaikuFirstRecourseOrderOne}{0.87}
\newcommand{\synHaikuFirstListed}{0.96}
\newcommand{\synHaikuFirstListedChance}{0.39}
\newcommand{\synHaikuSwitch}{0.92}
\newcommand{\synGemmaFourFirstValid}{0.92}

\newcommand{\synGemmaFourFirstTwoPlus}{0.51}

\newcommand{\synGemmaFourFirstRecourse}{0.85}

\newcommand{\synGemmaFourFirstFidelity}{0.64}

\newcommand{\synGemmaFourNeutralTwoPlus}{0.62}

\newcommand{\synGemmaFourNeutralRecourse}{0.85}

\newcommand{\synGemmaFourAllTwoPlus}{0.95}

\newcommand{\synGemmaFourAllRecourse}{0.99}

\newcommand{\synLlamaThreeFirstValid}{0.77}

\newcommand{\synLlamaThreeFirstTwoPlus}{0.15}

\newcommand{\synLlamaThreeFirstRecourse}{0.49}

\newcommand{\synLlamaThreeFirstFidelity}{0.46}

\newcommand{\synLlamaThreeNeutralTwoPlus}{0.22}

\newcommand{\synLlamaThreeNeutralRecourse}{0.55}

\newcommand{\synLlamaThreeAllTwoPlus}{0.45}

\newcommand{\synLlamaThreeAllRecourse}{0.70}

\newcommand{\synMistralFirstValid}{0.91}

\newcommand{\synMistralFirstTwoPlus}{0.05}

\newcommand{\synMistralFirstRecourse}{0.51}

\newcommand{\synMistralFirstFidelity}{0.61}

\newcommand{\synMistralNeutralTwoPlus}{0.17}

\newcommand{\synMistralNeutralRecourse}{0.53}

\newcommand{\synMistralAllTwoPlus}{0.86}

\newcommand{\synMistralAllRecourse}{0.95}

\newcommand{\synOlmoFirstValid}{0.72}

\newcommand{\synOlmoFirstTwoPlus}{0.14}

\newcommand{\synOlmoFirstRecourse}{0.48}

\newcommand{\synOlmoFirstFidelity}{0.29}

\newcommand{\synOlmoNeutralTwoPlus}{0.22}

\newcommand{\synOlmoNeutralRecourse}{0.52}

\newcommand{\synOlmoAllTwoPlus}{0.76}

\newcommand{\synOlmoAllRecourse}{0.86}

\newcommand{\synRecourseCIWidthMax}{0.14}
\newcommand{\synTwoPlusCIWidthMax}{0.13}

\newcommand{\synValidMinPct}{75\%}
\newcommand{\synValidMaxPct}{99\%}
\newcommand{\synValidMinModel}{Gemma-3-12B}
\newcommand{\synValidMaxModel}{Qwen2.5-14B}
\newcommand{\synRecourseMin}{0.49}
\newcommand{\synRecourseMax}{0.78}
\newcommand{\synRecourseMinModel}{Llama-3.1-8B}
\newcommand{\synRecourseMaxModel}{Qwen2.5-14B}

\newcommand{\synBlockedMinPct}{22\%}
\newcommand{\synBlockedMaxPct}{51\%}
\newcommand{\synTwoPlusMinPct}{14\%}
\newcommand{\synTwoPlusMaxPct}{69\%}
\newcommand{\synTwoPlusMinModel}{Llama-3.1-8B}
\newcommand{\synTwoPlusMaxModel}{Qwen2.5-14B}
\newcommand{\synAllRecourseMin}{0.90}
\newcommand{\synAllRecourseMax}{0.999}

\newcommand{\synAllTwoPlusMinPct}{71\%}
\newcommand{\synAllTwoPlusMaxPct}{97\%}

\newcommand{\synNeutralHighCount}{three}
\newcommand{\synNeutralHighMin}{0.80}
\newcommand{\synNeutralHighMax}{0.91}
\newcommand{\synFirstListedOpenMin}{0.36}
\newcommand{\synFirstListedOpenMax}{0.55}
\newcommand{\synFirstListedOpenMinModel}{Phi-4-14B}
\newcommand{\synFirstListedOpenMaxModel}{Qwen3-14B}
\newcommand{\synFidelityOpenMinPct}{36\%}
\newcommand{\synFidelityOpenMaxPct}{59\%}

\newcommand{\synAllOverNeutralHighMin}{0.06}
\newcommand{\synAllOverNeutralHighMax}{0.11}
\newcommand{\synAllOverNeutralLowMin}{0.33}
\newcommand{\synAllOverNeutralLowMax}{0.45}

\newcommand{\synBucketFourMinChange}{2.2}

\newcommand{\synBucketAllMin}{0.94}
\newcommand{\synAgreeGemmaTwelveLlamaEight}{0.77}
\newcommand{\synAgreeGemmaTwelveLlamaEightChance}{0.37}

\newcommand{\synAgreeGemmaTwelveQwenThree}{0.82}
\newcommand{\synAgreeGemmaTwelveQwenThreeChance}{0.43}

\newcommand{\synAgreeLlamaEightQwenThree}{0.73}
\newcommand{\synAgreeLlamaEightQwenThreeChance}{0.42}

\newcommand{\synAgreeFirstHaikuMin}{0.30}
\newcommand{\synAgreeFirstHaikuMax}{0.47}

\newcommand{\synTieAgreeNPairs}{ten}
\newcommand{\synTieAgreeMin}{0.60}
\newcommand{\synTieAgreeMax}{0.80}
\newcommand{\synTieAgreeChanceMin}{0.37}
\newcommand{\synTieAgreeChanceMax}{0.48}

\newcommand{\synSmallestMinPct}{45\%}
\newcommand{\synSmallestMaxPct}{73\%}
\newcommand{\synSmallestMinModel}{Llama-3.1-8B}
\newcommand{\synSmallestMaxModel}{Qwen2.5-14B}
\newcommand{\synSmallestUniformMinPct}{74\%}
\newcommand{\synSmallestUniformMaxPct}{86\%}

\newcommand{\realNCells}{1,172}
\newcommand{\realLlamaEightPolicyRecourse}{0.31}

\newcommand{\realLlamaEightPolicyPrecision}{0.75}
\newcommand{\realLlamaEightPolicyCited}{2.9}
\newcommand{\realLlamaEightPolicyFidelity}{0.75}

\newcommand{\realLlamaEightPolicyRecourseCI}{[0.27, 0.35]}

\newcommand{\realLlamaEightPolicyEligRecourse}{0.09}
\newcommand{\realLlamaEightPolicyEligActionable}{0.13}
\newcommand{\realLlamaEightPolicyAllRecourse}{0.55}

\newcommand{\realLlamaEightPolicyAllPrecision}{0.68}

\newcommand{\realLlamaEightFreeAllRecourse}{0.55}

\newcommand{\realLlamaEightFreeAllPrecision}{0.68}

\newcommand{\realLlamaEightOracleAllRecourse}{0.87}

\newcommand{\realLlamaEightOracleAllPrecision}{0.87}

\newcommand{\realLlamaEightPolicyAllNocapRecourse}{0.67}

\newcommand{\realLlamaEightPolicyAllNocapPrecision}{0.67}

\newcommand{\realLlamaEightCounterfactualRecourse}{0.29}

\newcommand{\realLlamaEightCounterfactualPrecision}{0.72}

\newcommand{\realLlamaEightMinhitRecourse}{0.58}

\newcommand{\realLlamaEightMinhitPrecision}{0.90}

\newcommand{\realLlamaEightRequesterRecourse}{0.53}

\newcommand{\realLlamaEightRequesterPrecision}{0.70}

\newcommand{\realLlamaEightRequesterCapRecourse}{0.28}

\newcommand{\realLlamaEightRequesterCapPrecision}{0.75}

\newcommand{\realGemmaTwelvePolicyRecourse}{0.23}

\newcommand{\realGemmaTwelvePolicyPrecision}{0.90}
\newcommand{\realGemmaTwelvePolicyCited}{2.2}
\newcommand{\realGemmaTwelvePolicyFidelity}{0.88}

\newcommand{\realGemmaTwelvePolicyRecourseCI}{[0.18, 0.28]}

\newcommand{\realGemmaTwelvePolicyEligRecourse}{0.07}
\newcommand{\realGemmaTwelvePolicyEligActionable}{0.13}
\newcommand{\realGemmaTwelvePolicyAllRecourse}{0.84}

\newcommand{\realGemmaTwelvePolicyAllPrecision}{0.75}

\newcommand{\realGemmaTwelveFreeAllRecourse}{0.81}

\newcommand{\realGemmaTwelveFreeAllPrecision}{0.69}

\newcommand{\realGemmaTwelveOracleAllRecourse}{0.997}

\newcommand{\realGemmaTwelveOracleAllPrecision}{0.86}

\newcommand{\realGemmaTwelvePolicyAllNocapRecourse}{0.84}

\newcommand{\realGemmaTwelvePolicyAllNocapPrecision}{0.71}

\newcommand{\realGemmaTwelveCounterfactualRecourse}{0.25}

\newcommand{\realGemmaTwelveCounterfactualPrecision}{0.93}

\newcommand{\realGemmaTwelveMinhitRecourse}{0.79}

\newcommand{\realGemmaTwelveMinhitPrecision}{0.96}

\newcommand{\realGemmaTwelveRequesterRecourse}{0.37}

\newcommand{\realGemmaTwelveRequesterPrecision}{0.82}

\newcommand{\realGemmaTwelveRequesterCapRecourse}{0.27}

\newcommand{\realGemmaTwelveRequesterCapPrecision}{0.90}

\newcommand{\realMistralPolicyRecourse}{0.24}

\newcommand{\realMistralPolicyPrecision}{0.81}
\newcommand{\realMistralPolicyCited}{2.5}
\newcommand{\realMistralPolicyFidelity}{0.74}

\newcommand{\realMistralPolicyRecourseCI}{[0.20, 0.29]}

\newcommand{\realMistralPolicyEligRecourse}{0.05}
\newcommand{\realMistralPolicyEligActionable}{0.08}
\newcommand{\realMistralPolicyAllRecourse}{0.58}

\newcommand{\realQwenThreePolicyRecourse}{0.40}

\newcommand{\realQwenThreePolicyPrecision}{0.88}
\newcommand{\realQwenThreePolicyCited}{3.1}
\newcommand{\realQwenThreePolicyFidelity}{0.92}

\newcommand{\realQwenThreePolicyRecourseCI}{[0.33, 0.46]}

\newcommand{\realQwenThreePolicyEligRecourse}{0.11}
\newcommand{\realQwenThreePolicyEligActionable}{0.17}
\newcommand{\realQwenThreePolicyAllRecourse}{0.79}

\newcommand{\realQwenThreePolicyAllPrecision}{0.89}

\newcommand{\realQwenThreeFreeAllRecourse}{0.71}

\newcommand{\realQwenThreeFreeAllPrecision}{0.75}

\newcommand{\realQwenThreeOracleAllRecourse}{0.98}

\newcommand{\realQwenThreeOracleAllPrecision}{0.98}

\newcommand{\realQwenThreePolicyAllNocapRecourse}{0.88}

\newcommand{\realQwenThreePolicyAllNocapPrecision}{0.75}

\newcommand{\realQwenThreeCounterfactualRecourse}{0.34}

\newcommand{\realQwenThreeCounterfactualPrecision}{0.92}

\newcommand{\realQwenThreeMinhitRecourse}{0.90}

\newcommand{\realQwenThreeMinhitPrecision}{0.96}

\newcommand{\realQwenThreeRequesterRecourse}{0.59}

\newcommand{\realQwenThreeRequesterPrecision}{0.75}

\newcommand{\realQwenThreeRequesterCapRecourse}{0.36}

\newcommand{\realQwenThreeRequesterCapPrecision}{0.94}

\newcommand{\realQwenTwoFivePolicyRecourse}{0.51}

\newcommand{\realQwenTwoFivePolicyPrecision}{0.91}
\newcommand{\realQwenTwoFivePolicyCited}{3.2}
\newcommand{\realQwenTwoFivePolicyFidelity}{0.91}

\newcommand{\realQwenTwoFivePolicyRecourseCI}{[0.45, 0.57]}

\newcommand{\realQwenTwoFivePolicyEligRecourse}{0.16}
\newcommand{\realQwenTwoFivePolicyEligActionable}{0.24}
\newcommand{\realQwenTwoFivePolicyAllRecourse}{0.77}

\newcommand{\realPhiFourPolicyRecourse}{0.57}

\newcommand{\realPhiFourPolicyPrecision}{0.82}
\newcommand{\realPhiFourPolicyCited}{3.9}
\newcommand{\realPhiFourPolicyFidelity}{0.93}

\newcommand{\realPhiFourPolicyRecourseCI}{[0.52, 0.63]}

\newcommand{\realPhiFourPolicyEligRecourse}{0.36}
\newcommand{\realPhiFourPolicyEligActionable}{0.44}
\newcommand{\realPhiFourPolicyAllRecourse}{0.72}

\newcommand{\realQwenThreeFivePolicyRecourse}{0.47}

\newcommand{\realQwenThreeFivePolicyPrecision}{0.80}
\newcommand{\realQwenThreeFivePolicyCited}{3.5}
\newcommand{\realQwenThreeFivePolicyFidelity}{0.87}

\newcommand{\realQwenThreeFivePolicyRecourseCI}{[0.42, 0.52]}

\newcommand{\realQwenThreeFivePolicyEligRecourse}{0.35}
\newcommand{\realQwenThreeFivePolicyEligActionable}{0.39}
\newcommand{\realQwenThreeFivePolicyAllRecourse}{0.80}

\newcommand{\realGemmaNewPolicyRecourse}{0.70}

\newcommand{\realGemmaNewPolicyPrecision}{0.95}
\newcommand{\realGemmaNewPolicyCited}{3.0}
\newcommand{\realGemmaNewPolicyFidelity}{0.82}

\newcommand{\realGemmaNewPolicyRecourseCI}{[0.65, 0.76]}

\newcommand{\realGemmaNewPolicyEligRecourse}{0.50}
\newcommand{\realGemmaNewPolicyEligActionable}{0.54}
\newcommand{\realGemmaNewPolicyAllRecourse}{0.84}

\newcommand{\realMinistralPolicyRecourse}{0.67}

\newcommand{\realMinistralPolicyPrecision}{0.79}
\newcommand{\realMinistralPolicyCited}{4.1}
\newcommand{\realMinistralPolicyFidelity}{0.92}

\newcommand{\realMinistralPolicyRecourseCI}{[0.61, 0.72]}

\newcommand{\realMinistralPolicyEligRecourse}{0.44}
\newcommand{\realMinistralPolicyEligActionable}{0.55}
\newcommand{\realMinistralPolicyAllRecourse}{0.91}

\newcommand{\realGranitePolicyRecourse}{0.65}

\newcommand{\realGranitePolicyPrecision}{0.76}
\newcommand{\realGranitePolicyCited}{4.7}
\newcommand{\realGranitePolicyFidelity}{0.96}

\newcommand{\realGranitePolicyRecourseCI}{[0.59, 0.70]}

\newcommand{\realGranitePolicyEligRecourse}{0.37}
\newcommand{\realGranitePolicyEligActionable}{0.46}
\newcommand{\realGranitePolicyAllRecourse}{0.88}

\newcommand{\realGptOssPolicyRecourse}{0.55}

\newcommand{\realGptOssPolicyPrecision}{0.91}
\newcommand{\realGptOssPolicyCited}{3.0}
\newcommand{\realGptOssPolicyFidelity}{0.87}

\newcommand{\realGptOssPolicyRecourseCI}{[0.50, 0.60]}

\newcommand{\realGptOssPolicyEligRecourse}{0.44}
\newcommand{\realGptOssPolicyEligActionable}{0.50}
\newcommand{\realGptOssPolicyAllRecourse}{0.90}

\newcommand{\realGemmaTwentySevenPolicyRecourse}{0.44}

\newcommand{\realGemmaTwentySevenPolicyPrecision}{0.85}
\newcommand{\realGemmaTwentySevenPolicyCited}{3.2}
\newcommand{\realGemmaTwentySevenPolicyFidelity}{0.90}

\newcommand{\realGemmaTwentySevenPolicyRecourseCI}{[0.38, 0.50]}

\newcommand{\realGemmaTwentySevenPolicyEligRecourse}{0.40}
\newcommand{\realGemmaTwentySevenPolicyEligActionable}{0.53}
\newcommand{\realGemmaTwentySevenPolicyAllRecourse}{0.87}

\newcommand{\realQwenThreeBigPolicyRecourse}{0.54}

\newcommand{\realQwenThreeBigPolicyPrecision}{0.90}
\newcommand{\realQwenThreeBigPolicyCited}{3.3}
\newcommand{\realQwenThreeBigPolicyFidelity}{0.87}

\newcommand{\realQwenThreeBigPolicyRecourseCI}{[0.48, 0.59]}

\newcommand{\realQwenThreeBigPolicyEligRecourse}{0.32}
\newcommand{\realQwenThreeBigPolicyEligActionable}{0.42}
\newcommand{\realQwenThreeBigPolicyAllRecourse}{0.84}

\newcommand{\realQwenThreeEightPolicyRecourse}{0.77}

\newcommand{\realQwenThreeEightPolicyPrecision}{0.94}
\newcommand{\realQwenThreeEightPolicyCited}{3.3}
\newcommand{\realQwenThreeEightPolicyFidelity}{0.83}

\newcommand{\realQwenThreeEightPolicyRecourseCI}{[0.72, 0.81]}

\newcommand{\realQwenThreeEightPolicyEligRecourse}{0.74}
\newcommand{\realQwenThreeEightPolicyEligActionable}{0.84}
\newcommand{\realQwenThreeEightPolicyAllRecourse}{0.93}

\newcommand{\realGemmaNewBigPolicyRecourse}{0.97}

\newcommand{\realGemmaNewBigPolicyPrecision}{0.93}
\newcommand{\realGemmaNewBigPolicyCited}{3.5}
\newcommand{\realGemmaNewBigPolicyFidelity}{0.80}

\newcommand{\realGemmaNewBigPolicyRecourseCI}{[0.95, 0.99]}

\newcommand{\realGemmaNewBigPolicyHudRecourse}{1.00}

\newcommand{\realGemmaNewBigPolicyEligRecourse}{0.97}
\newcommand{\realGemmaNewBigPolicyEligActionable}{0.99}
\newcommand{\realGemmaNewBigPolicyAllRecourse}{0.98}

\newcommand{\realGemmaNewBigFreeRecourse}{0.58}

\newcommand{\realGemmaNewBigFreeHudRecourse}{0.15}

\newcommand{\realHaikuPolicyRecourse}{0.96}

\newcommand{\realHaikuPolicyPrecision}{0.93}
\newcommand{\realHaikuPolicyCited}{3.9}
\newcommand{\realHaikuPolicyFidelity}{0.89}

\newcommand{\realHaikuPolicyRecourseCI}{[0.93, 0.98]}

\newcommand{\realHaikuPolicyEligRecourse}{0.90}
\newcommand{\realHaikuPolicyEligActionable}{0.96}
\newcommand{\realHaikuPolicyAllRecourse}{0.98}

\newcommand{\realLunaPolicyRecourse}{0.92}

\newcommand{\realLunaPolicyPrecision}{0.89}
\newcommand{\realLunaPolicyCited}{3.8}
\newcommand{\realLunaPolicyFidelity}{0.84}

\newcommand{\realLunaPolicyRecourseCI}{[0.89, 0.94]}

\newcommand{\realLunaPolicyEligRecourse}{0.87}
\newcommand{\realLunaPolicyEligActionable}{0.98}
\newcommand{\realLunaPolicyAllRecourse}{0.97}

\newcommand{\realSolPolicyRecourse}{0.99}

\newcommand{\realSolPolicyPrecision}{0.83}
\newcommand{\realSolPolicyCited}{4.4}
\newcommand{\realSolPolicyFidelity}{0.85}

\newcommand{\realSolPolicyRecourseCI}{[0.98, 1.00]}

\newcommand{\realSolPolicyEligRecourse}{0.99}
\newcommand{\realSolPolicyEligActionable}{1.00}
\newcommand{\realSolPolicyAllRecourse}{0.99}

\newcommand{\realSonnetPolicyRecourse}{0.94}

\newcommand{\realSonnetPolicyPrecision}{0.82}
\newcommand{\realSonnetPolicyCited}{4.9}
\newcommand{\realSonnetPolicyFidelity}{0.97}

\newcommand{\realSonnetPolicyRecourseCI}{[0.92, 0.97]}

\newcommand{\realSonnetPolicyEligRecourse}{0.87}
\newcommand{\realSonnetPolicyEligActionable}{0.96}
\newcommand{\realSonnetPolicyAllRecourse}{0.96}

\newcommand{\realOpusPolicyRecourse}{0.99}

\newcommand{\realOpusPolicyPrecision}{0.81}
\newcommand{\realOpusPolicyCited}{4.9}
\newcommand{\realOpusPolicyFidelity}{0.96}

\newcommand{\realOpusPolicyRecourseCI}{[0.98, 1.00]}

\newcommand{\realOpusPolicyEligRecourse}{0.99}
\newcommand{\realOpusPolicyEligActionable}{1.00}
\newcommand{\realOpusPolicyAllRecourse}{0.99}

\newcommand{\realNModels}{11}

\newcommand{\realNModelsAblate}{three}
\newcommand{\realMeanPolicyRecourse}{0.48}
\newcommand{\realMeanPolicyTwoPlus}{0.43}

\newcommand{\realMeanPolicyCited}{3.3}
\newcommand{\realMeanPolicyFidelity}{0.87}

\newcommand{\realMeanPolicyControlRecourse}{0.88}
\newcommand{\realMeanPolicyDocCloudRecourse}{0.71}
\newcommand{\realMeanPolicyDocCloudTwoPlus}{0.37}

\newcommand{\realMeanPolicyDocCloudFidelity}{0.94}

\newcommand{\realMeanPolicyDocCloudControlRecourse}{0.97}
\newcommand{\realMeanPolicyGithubRecourse}{0.91}
\newcommand{\realMeanPolicyGithubTwoPlus}{0.19}

\newcommand{\realMeanPolicyGithubFidelity}{0.58}

\newcommand{\realMeanPolicyGithubControlRecourse}{0.92}
\newcommand{\realMeanPolicyReposRecourse}{0.49}
\newcommand{\realMeanPolicyReposTwoPlus}{0.29}

\newcommand{\realMeanPolicyReposFidelity}{0.87}

\newcommand{\realMeanPolicyReposControlRecourse}{0.91}
\newcommand{\realMeanPolicyPrivilegedRecourse}{0.50}
\newcommand{\realMeanPolicyPrivilegedTwoPlus}{0.23}

\newcommand{\realMeanPolicyPrivilegedFidelity}{0.86}

\newcommand{\realMeanPolicyPrivilegedControlRecourse}{0.81}
\newcommand{\realMeanPolicySnapRecourse}{0.41}
\newcommand{\realMeanPolicySnapTwoPlus}{0.61}

\newcommand{\realMeanPolicySnapFidelity}{0.96}

\newcommand{\realMeanPolicySnapControlRecourse}{0.85}
\newcommand{\realMeanPolicyTitleFourRecourse}{0.19}
\newcommand{\realMeanPolicyTitleFourTwoPlus}{0.58}

\newcommand{\realMeanPolicyTitleFourFidelity}{0.89}

\newcommand{\realMeanPolicyTitleFourControlRecourse}{0.84}
\newcommand{\realMeanPolicyHudRecourse}{0.20}
\newcommand{\realMeanPolicyHudTwoPlus}{0.76}

\newcommand{\realMeanPolicyHudFidelity}{0.99}

\newcommand{\realMeanPolicyHudControlRecourse}{0.91}

\newcommand{\realMeanPolicyAllRecourse}{0.78}

\newcommand{\realMeanPolicyAllCited}{4.6}

\newcommand{\realMeanPolicyAllDocCloudRecourse}{0.92}

\newcommand{\realMeanPolicyAllGithubRecourse}{0.98}

\newcommand{\realMeanPolicyAllReposRecourse}{0.84}

\newcommand{\realMeanPolicyAllPrivilegedRecourse}{0.92}

\newcommand{\realMeanPolicyAllSnapRecourse}{0.78}

\newcommand{\realMeanPolicyAllTitleFourRecourse}{0.47}

\newcommand{\realMeanPolicyAllHudRecourse}{0.58}

\newcommand{\realMeanOracleAllRecourse}{0.95}

\newcommand{\realMinDefault}{0.23}
\newcommand{\realMaxDefault}{0.70}

\newcommand{\realMinAll}{0.55}
\newcommand{\realMaxAll}{0.91}

\newcommand{\realNPrecisionFall}{ten}

\newcommand{\realBucketOnePolicy}{0.89}
\newcommand{\realBucketOneN}{221}
\newcommand{\realBucketOneMinChange}{1.0}
\newcommand{\realBucketTwoPolicy}{0.59}
\newcommand{\realBucketTwoN}{102}
\newcommand{\realBucketTwoMinChange}{1.9}
\newcommand{\realBucketThreePolicy}{0.33}
\newcommand{\realBucketThreeN}{50}
\newcommand{\realBucketThreeMinChange}{2.9}
\newcommand{\realBucketFourPolicy}{0.12}
\newcommand{\realBucketFourN}{34}
\newcommand{\realBucketFourMinChange}{4.5}
\newcommand{\realBucketOnePolicyAll}{0.96}
\newcommand{\realBucketTwoPolicyAll}{0.90}
\newcommand{\realBucketThreePolicyAll}{0.71}
\newcommand{\realBucketFourPolicyAll}{0.43}
\newcommand{\realLlamaEightPolicyRandom}{0.24}
\newcommand{\realGemmaTwelvePolicyRandom}{0.18}
\newcommand{\realMistralPolicyRandom}{0.20}
\newcommand{\realQwenThreePolicyRandom}{0.27}
\newcommand{\realQwenTwoFivePolicyRandom}{0.29}
\newcommand{\realPhiFourPolicyRandom}{0.34}
\newcommand{\realQwenThreeFivePolicyRandom}{0.26}
\newcommand{\realGemmaNewPolicyRandom}{0.24}
\newcommand{\realMinistralPolicyRandom}{0.36}
\newcommand{\realGranitePolicyRandom}{0.43}
\newcommand{\realGptOssPolicyRandom}{0.23}
\newcommand{\realGemmaTwentySevenPolicyRandom}{0.23}
\newcommand{\realQwenThreeBigPolicyRandom}{0.28}
\newcommand{\realQwenThreeEightPolicyRandom}{0.25}
\newcommand{\realGemmaNewBigPolicyRandom}{0.28}
\newcommand{\realHaikuPolicyRandom}{0.33}
\newcommand{\realLunaPolicyRandom}{0.33}
\newcommand{\realSolPolicyRandom}{0.39}
\newcommand{\realSonnetPolicyRandom}{0.45}
\newcommand{\realOpusPolicyRandom}{0.45}

\newcommand{\realFreeFidelityDropMin}{0.11}
\newcommand{\realFreeFidelityDropMax}{0.30}

\newcommand{\realOracleAllOverMinhitMin}{0.08}
\newcommand{\realOracleAllOverMinhitMax}{0.29}

\newcommand{\realCounterfactualOverPolicyMin}{$-$0.06}
\newcommand{\realCounterfactualOverPolicyMax}{0.02}

\newcommand{\realPolicyAllNocapOverPolicyAllMin}{0.003}
\newcommand{\realPolicyAllNocapOverPolicyAllMax}{0.12}

\newcommand{\realMinhitMissMinPct}{2\%}
\newcommand{\realMinhitMissMaxPct}{8\%}
\newcommand{\realNGens}{152,754}
\newcommand{\realDropped}{203}
\newcommand{\realAgreeAlpha}{0.90}
\newcommand{\realAgreeMicroFOne}{0.94}

\newcommand{\realAgreeN}{2,000}
\newcommand{\realAgreeModel}{Gemma-3-12B}
\newcommand{\synProbeGemmaTwelveREight}{0.38}
\newcommand{\synProbeGemmaTwelveRSeven}{0.26}
\newcommand{\synProbeGemmaTwelveN}{221}
\newcommand{\synProbeLlamaEightREight}{0.23}
\newcommand{\synProbeLlamaEightRSeven}{0.28}
\newcommand{\synProbeLlamaEightN}{237}
\newcommand{\synProbeQwenThreeREight}{0.54}
\newcommand{\synProbeQwenThreeRSeven}{0.14}
\newcommand{\synProbeQwenThreeN}{98}

\newcommand{\realHumanN}{200}

\newcommand{\realHumanXAlpha}{0.83}
\newcommand{\realNSmall}{11}
\newcommand{\realNLarge}{four}
\newcommand{\realNFrontier}{five}
\newcommand{\realNModelsAll}{20}
\newcommand{\realLargeMinDefault}{0.44}
\newcommand{\realLargeMaxDefault}{0.97}

\newcommand{\realFrontierMinDefault}{0.92}
\newcommand{\realFrontierMaxDefault}{0.99}

\newcommand{\realCapEciN}{15}
\newcommand{\realCapEciRho}{0.88}

\newcommand{\realCapFidN}{20}
\newcommand{\realCapFidRho}{0.14}

\newcommand{\realCapFidMin}{0.74}
\newcommand{\realCapFidMax}{0.97}
\newcommand{\realCapBandLo}{138}
\newcommand{\realCapBandHi}{143}
\newcommand{\realCapBandMin}{0.40}
\newcommand{\realCapBandMax}{0.97}
\newcommand{\realCapBandN}{five}
\newcommand{\realRecallGapN}{17,859}
\newcommand{\realRecallGapPct}{37\%}
\newcommand{\realRecallGapLlamaEightPct}{66\%}
\newcommand{\realRecallGapGemmaTwelvePct}{79\%}
\newcommand{\realRecallGapMistralPct}{78\%}
\newcommand{\realRecallGapQwenThreePct}{63\%}
\newcommand{\realRecallGapQwenTwoFivePct}{50\%}
\newcommand{\realRecallGapPhiFourPct}{44\%}
\newcommand{\realRecallGapQwenThreeFivePct}{57\%}
\newcommand{\realRecallGapGemmaNewPct}{34\%}
\newcommand{\realRecallGapMinistralPct}{33\%}
\newcommand{\realRecallGapGranitePct}{36\%}
\newcommand{\realRecallGapGptOssPct}{45\%}
\newcommand{\realRecallGapGemmaTwentySevenPct}{59\%}
\newcommand{\realRecallGapQwenThreeBigPct}{50\%}
\newcommand{\realRecallGapQwenThreeEightPct}{26\%}
\newcommand{\realRecallGapGemmaNewBigPct}{2\%}
\newcommand{\realRecallGapHaikuPct}{5\%}
\newcommand{\realRecallGapLunaPct}{9\%}
\newcommand{\realRecallGapSolPct}{1\%}
\newcommand{\realRecallGapSonnetPct}{6\%}
\newcommand{\realRecallGapOpusPct}{1\%}

\newcommand{\realRecallGapSmallMaxPct}{79\%}

\newcommand{\realRecallGapSmallMaxModel}{Gemma-3-12B}

\newcommand{\realJudgeLunaJudgeN}{2,800}
\newcommand{\realJudgeLunaJudgeNModels}{20}
\newcommand{\realJudgeLunaJudgeAurocCorrect}{0.62}

\newcommand{\realJudgeLunaJudgeAurocActionable}{0.97}

\newcommand{\realJudgeLunaJudgeRhoActionable}{0.99}
\newcommand{\realJudgeLunaJudgeAurocQuality}{0.95}

\newcommand{\realJudgeLunaJudgeCorrectHighPct}{84\%}

\newcommand{\realJudgeGemmaJudgeAurocCorrect}{0.57}

\newcommand{\realJudgeGemmaJudgeAurocActionable}{0.92}

\newcommand{\realJudgeGemmaJudgeRhoActionable}{0.98}
\newcommand{\realJudgeGemmaJudgeAurocQuality}{0.93}

\newcommand{\realJudgeGemmaJudgeCorrectHighPct}{87\%}

\newcommand{\realJudgeQwenJudgeAurocCorrect}{0.56}

\newcommand{\realJudgeQwenJudgeAurocActionable}{0.67}

\newcommand{\realJudgeQwenJudgeAurocQuality}{0.84}

\newcommand{\realJudgeQwenJudgeCorrectHighPct}{89\%}
\newcommand{\realJudgeAurocCorrectMin}{0.56}
\newcommand{\realJudgeAurocCorrectMax}{0.62}

\newcommand{\realJudgeAurocCorrectWithinMin}{0.51}
\newcommand{\realJudgeAurocCorrectWithinMax}{0.55}

\newcommand{\realJudgeCorrectFidMin}{0.54}
\newcommand{\realJudgeCorrectFidMax}{0.56}

\newcommand{\realJudgeCorrectHighPctMin}{84\%}
\newcommand{\realJudgeCorrectHighPctMax}{89\%}

\newcommand{\realJudgeNJudges}{three}
\newcommand{\realListAllCeiling}{0.90}
\newcommand{\realListAllDeltaMin}{0.14}
\newcommand{\realListAllDeltaMax}{0.61}
\newcommand{\realListAllDeltaMinModel}{Gemma-4-12B}
\newcommand{\realListAllDeltaMaxModel}{Gemma-3-12B}

\newcommand{\realListAllDeltaFrontierMax}{0.05}
\newcommand{\realListAllNOpen}{14}
\newcommand{\realHumanMicroFOne}{0.91}
\newcommand{\realHumanRecourseDiffMax}{0.06}
\newcommand{\realAgreeNewAlpha}{0.83}

\newcommand{\realAgreeNewN}{1,398}
\newcommand{\realAgreeNewNModels}{14}
\newcommand{\realAgreeNewRecourseDiffMax}{0.04}

\newcommand{\realAgreeNModelsCovered}{20}
\newcommand{\realAurocN}{20,375}

\newcommand{\realAurocFidelity}{0.49}
\newcommand{\realAurocFidelityCI}{[0.47, 0.50]}
\newcommand{\realAurocFidelityWithin}{0.48}
\newcommand{\realAurocRecall}{0.48}
\newcommand{\realAurocRecallCI}{[0.47, 0.50]}
\newcommand{\realAurocRecallWithin}{0.47}
\newcommand{\realAurocPrecision}{0.48}
\newcommand{\realAurocPrecisionCI}{[0.46, 0.50]}
\newcommand{\realAurocPrecisionWithin}{0.45}
\newcommand{\realAurocCoverage}{0.73}
\newcommand{\realAurocCoverageCI}{[0.70, 0.75]}
\newcommand{\realAurocCoverageWithin}{0.71}
\newcommand{\realAurocTwoPlus}{0.60}
\newcommand{\realAurocTwoPlusCI}{[0.58, 0.63]}
\newcommand{\realAurocTwoPlusWithin}{0.51}
\newcommand{\realAurocCited}{0.75}
\newcommand{\realAurocCitedCI}{[0.73, 0.78]}
\newcommand{\realAurocCitedWithin}{0.67}
\newcommand{\realCoveredZeroRV}{0.48}

\newcommand{\realCoveredOneRV}{0.49}

\newcommand{\realCoveredTwoRV}{0.75}

\newcommand{\realValidIncompletePct}{23\%}
\newcommand{\realCapEciOpenN}{10}
\newcommand{\realCapEciOpenRho}{0.71}

\newcommand{\realEligSmallTitleFourA}{0.26}

\newcommand{\realEligSmallHudA}{0.36}

\newcommand{\realEligLargeTitleFourA}{0.60}

\newcommand{\realEligLargeHudA}{0.69}

\newcommand{\realEligFrontierTitleFourA}{0.96}

\newcommand{\realEligFrontierHudA}{0.98}
\newcommand{\realAllMarginMin}{0.04}
\newcommand{\realAllMarginMax}{0.70}
\newcommand{\realAllMarginMinModel}{Mistral-Nemo-12B}
\newcommand{\realAllMarginMaxModel}{Gemma-4-31B}

\newcommand{\realCiTouchRandomModels}{Gemma-3-12B and Mistral-Nemo-12B}
\newcommand{\realListAllGapShareMin}{35\%}
\newcommand{\realListAllGapShareMax}{79\%}
\newcommand{\realListAllGapShareMinModel}{Phi-4-14B}
\newcommand{\realListAllGapShareMaxModel}{Gemma-3-12B}
\newcommand{\realListAllNUnderHalf}{four}
\newcommand{\realFreeOverPolicyMin}{$-$0.19}
\newcommand{\realFreeOverPolicyMax}{0.29}
\newcommand{\realFreeOverPolicyMinModel}{Gemma-4-12B}
\newcommand{\realFreeOverPolicyMaxModel}{Gemma-3-12B}
\newcommand{\realStratumMultiNone}{6}
\newcommand{\realStratumMultiOne}{18}
\newcommand{\realStratumMultiTwo}{186}
\newcommand{\realStratumSingleNone}{7}

\newcommand{\realStratumDefined}{407}
\newcommand{\realGptOssAtBudgetPct}{0\%}
\newcommand{\realGptOssMaxTokens}{458}
\newcommand{\realCapHitMaxPct}{13\%}
\newcommand{\realCapHitMaxModel}{Granite-4.2-8B}
\newcommand{\realThinkOffMaxTokens}{180}

\newcommand{\realCapHitRestMaxPct}{1\%}

\newcommand{\realNThink}{four}
\newcommand{\realThinkOnMin}{0.89}
\newcommand{\realThinkOnMax}{0.99}
\newcommand{\realThinkOffMin}{0.40}
\newcommand{\realThinkOffMax}{0.77}

\newcommand{\realThinkCitedGainMin}{0.5}
\newcommand{\realThinkCitedGainMax}{0.9}
\newcommand{\realThinkPrecDiffMin}{$-$0.04}
\newcommand{\realThinkPrecDiffMax}{0.07}
\newcommand{\realQEightDiffMin}{$-$0.06}
\newcommand{\realQEightDiffMax}{0.09}
\newcommand{\realRequesterNModels}{13}

\newcommand{\realRequesterNSmall}{11}
\newcommand{\realRequesterNRaise}{nine}
\newcommand{\realRequesterNBelowAll}{ten}
\newcommand{\realRequesterMeanGain}{0.14}
\newcommand{\realRequesterMeanAllGap}{0.16}
\newcommand{\realRequesterFrontierDiffMax}{0.002}
\newcommand{\realRequesterFrontierModels}{Claude Haiku 4.5 and GPT-6 Luna}
\newcommand{\realRequesterCapOverPolicyMin}{$-$0.03}
\newcommand{\realRequesterCapOverPolicyMax}{0.04}
\newcommand{\realRequesterOverCapMin}{0.10}
\newcommand{\realRequesterOverCapMax}{0.25}
\newcommand{\realExecMinhitOutsideN}{83}

\newcommand{\realExecCounterfactualAdviceMin}{0.81}
\newcommand{\realExecCounterfactualAdviceMax}{0.86}

\newcommand{\realExecCounterfactualExecGivenRVMin}{0.59}
\newcommand{\realExecCounterfactualExecGivenRVMax}{0.64}

\newcommand{\realExecCounterfactualExecGivenNotRVMax}{0.02}

\newcommand{\realExecCounterfactualExecRVMin}{0.25}
\newcommand{\realExecCounterfactualExecRVMax}{0.34}

\newcommand{\realExecCounterfactualExecRVDefinedMin}{0.17}
\newcommand{\realExecCounterfactualExecRVDefinedMax}{0.21}

\newcommand{\realExecMinhitExecGivenNotRVMax}{0.24}

\newcommand{\realExecMinhitExecRVMin}{0.58}
\newcommand{\realExecMinhitExecRVMax}{0.90}

\newcommand{\realExecMinhitExecRVDefinedMin}{0.30}
\newcommand{\realExecMinhitExecRVDefinedMax}{0.69}

\newcommand{\realExecRequesterExecutedMin}{0.08}
\newcommand{\realExecRequesterExecutedMax}{0.56}

\newcommand{\realExecRequesterExecRVMin}{0.37}
\newcommand{\realExecRequesterExecRVMax}{0.95}

\newcommand{\oracleDocCloudNoSingle}{0.95}
\newcommand{\oracleGithubNoSingle}{0.00}
\newcommand{\oracleReposNoSingle}{1.00}
\newcommand{\oraclePrivilegedNoSingle}{1.00}
\newcommand{\oracleSnapNoSingle}{0.95}
\newcommand{\oracleTitleFourNoSingle}{0.998}
\newcommand{\oracleHudNoSingle}{1.00}
\newcommand{\oracleNoSingleMin}{0.95}
\newcommand{\oracleNoSingleMax}{1.00}

\newcommand{\realBucketOneFrontierPolicy}{1.00}
\newcommand{\realBucketTwoFrontierPolicy}{0.99}
\newcommand{\realBucketThreeFrontierPolicy}{0.98}
\newcommand{\realBucketFourFrontierPolicy}{0.84}
\newcommand{\realBucketFourFrontierMin}{0.76}

\newcommand{\realBucketOneThinkMean}{1.00}
\newcommand{\realBucketTwoThinkMean}{0.97}
\newcommand{\realBucketThreeThinkMean}{0.92}
\newcommand{\realBucketFourThinkMean}{0.76}
\newcommand{\realBucketFourThinkMin}{0.62}
\newcommand{\realBucketFourThinkMax}{0.94}
\newcommand{\realBucketTwoThinkMin}{0.94}
\newcommand{\realBucketTwoThinkMax}{1.00}
\newcommand{\realFourNBelowEight}{six}
\newcommand{\realFourNReasoning}{nine}
\newcommand{\realHighFourHaiku}{0.78}

\newcommand{\realHighFourLuna}{0.88}

\newcommand{\realHighFourSonnet}{0.93}
\newcommand{\realHighFourSonnetCI}{[0.86, 0.98]}

\newcommand{\realThinkAllMin}{0.97}
\newcommand{\realThinkAllMax}{0.98}
\newcommand{\realThinkAllFourMin}{0.84}
\newcommand{\realThinkAllFourMax}{0.90}
\newcommand{\realThinkAllBaseFourMin}{0.62}
\newcommand{\realThinkAllBaseFourMax}{0.79}
\newcommand{\realThinkAurocFidelity}{0.49}
\newcommand{\realThinkAurocFidelityCI}{[0.46, 0.52]}
\newcommand{\realThinkAurocRecall}{0.48}

\newcommand{\realThinkAurocCoverage}{0.75}

\newcommand{\realThinkAurocN}{4,072}
\newcommand{\realThinkShareRV}{93\%}
\newcommand{\realAurocSingleFidelity}{0.93}
\newcommand{\realAurocSingleFidelityCI}{[0.91, 0.94]}

\newcommand{\realAurocMultiFidelity}{0.49}
\newcommand{\realAurocMultiFidelityCI}{[0.48, 0.51]}

\newcommand{\realAurocFidelityGithub}{0.78}
\newcommand{\realAurocFidelityGithubCI}{[0.75, 0.80]}
\newcommand{\realAurocFidelityOtherMin}{0.50}
\newcommand{\realAurocFidelityOtherMax}{0.57}
\newcommand{\realAurocFidelityWithinEp}{0.56}
\newcommand{\realAurocRecallWithinEp}{0.56}
\newcommand{\realAurocPrecisionWithinEp}{0.58}
\newcommand{\realAurocCoverageWithinEp}{0.83}
\newcommand{\realAurocTwoPlusWithinEp}{0.70}
\newcommand{\realAurocCitedWithinEp}{0.87}
\newcommand{\realAurocWithinEpN}{202}
\newcommand{\realAurocFidelityWithinMin}{0.42}
\newcommand{\realAurocFidelityWithinMax}{0.64}
\newcommand{\realLogitFidelityCoef}{$-$0.46}

\newcommand{\realLogitFidelityCI}{[$-$2.55, 1.63]}

\newcommand{\realLogitCoverageP}{$p<0.001$}

\newcommand{\realReasonNullNSmallBelow}{11}

\newcommand{\realReasonNullSmallMinModel}{Llama-3.1-8B}
\newcommand{\realReasonNullFrontierMin}{$-$0.001}
\newcommand{\realReasonNullFrontierMax}{0.05}

\newcommand{\realReasonNullSmallGapMax}{0.26}
\newcommand{\realCapEciCI}{[0.64, 0.96]}
\newcommand{\realCapEciOpenCI}{[0.06, 0.92]}
\newcommand{\realCapFidCI}{[$-$0.38, 0.59]}

\newcommand{\realCapEciSmallN}{seven}
\newcommand{\realCapEciSmallRho}{0.54}
\newcommand{\realCapEciSmallP}{$p=0.23$}
\newcommand{\realEciGemmaTwelve}{123.5}
\newcommand{\realEciGemmaTwentySeven}{130.0}
\newcommand{\realEciGemmaNewBig}{142.7}
\newcommand{\realEciLuna}{156.3}
\newcommand{\realEciSol}{166.1}
\newcommand{\realEciGptOss}{137.8}
\newcommand{\realEciHaiku}{142.4}
\newcommand{\realEciLlamaEight}{116.6}
\newcommand{\realEciMistral}{118.7}
\newcommand{\realEciOpus}{167.3}
\newcommand{\realEciPhiFour}{130.4}
\newcommand{\realEciQwenThree}{138.2}
\newcommand{\realEciQwenThreeBig}{138.5}
\newcommand{\realEciQwenThreeFive}{139.5}
\newcommand{\realEciSonnet}{165.0}
\newcommand{\realControlTwoPlusN}{13}
\newcommand{\realControlTwoPlusSmallMean}{0.87}

\newcommand{\realEligSnapNActionable}{13}
\newcommand{\realEligTitleFourNActionable}{13}
\newcommand{\realEligHudNActionable}{7}

\newcommand{\realJudgeLunaJudgeAurocActionableCI}{[0.96, 0.98]}

\newcommand{\realJudgeLunaJudgeActMinusQual}{0.02}

\newcommand{\realJudgeGemmaJudgeAurocActionableCI}{[0.90, 0.94]}

\newcommand{\realJudgeGemmaJudgeActMinusQual}{$-$0.01}

\newcommand{\realJudgeQwenJudgeActMinusQual}{$-$0.16}

\newcommand{\realJudgeSelfShiftMax}{0.005}
\newcommand{\realJudgeOnlyActionableN}{700}
\newcommand{\realJudgeOnlyActionableAlone}{0.97}
\newcommand{\realJudgeOnlyActionableAloneCI}{[0.95, 0.98]}
\newcommand{\realJudgeOnlyActionableInThree}{0.96}

\newcommand{\realJudgeOnlyQualityAlone}{0.89}
\newcommand{\realJudgeOnlyQualityAloneCI}{[0.86, 0.91]}
\newcommand{\realJudgeOnlyQualityInThree}{0.94}

\newcommand{\realHumanVerdictAgreePct}{95\%}
\newcommand{\realHumanKappa}{0.91}
\newcommand{\realHumanKappaCI}{[0.84, 0.96]}
\newcommand{\realHumanExtractorOnly}{eight}
\newcommand{\realHumanHumanOnly}{one}
\newcommand{\realNGensPaper}{141,750}
\newcommand{\realNGensSmall}{118,650}
\newcommand{\realNGensLarge}{10,500}
\newcommand{\realNGensFrontier}{12,600}

\newcommand{\simReaders}{Gemma-3-12B, Llama-3.1-8B, Mistral-Nemo-12B, Phi-4-14B, Qwen2.5-14B and Qwen3-14B}

\newcommand{\simOneUnblockMin}{0.01}
\newcommand{\simOneUnblockMax}{0.04}

\newcommand{\simAllUnblockMin}{0.11}
\newcommand{\simAllUnblockMax}{0.96}

\newcommand{\simOneConfined}{0.73}
\newcommand{\simHOneDiffMin}{0.10}
\newcommand{\simHOneDiffMax}{0.93}
\newcommand{\simHOneNSig}{six}

\newcommand{\allow}{\textup{\textsc{allow}}\xspace}
\newcommand{\deny}{\textup{\textsc{deny}}\xspace}
\newcommand{\RV}{\mathrm{RV}}
\newcommand{\MSR}{\mathrm{MSR}}
\newcommand{\feat}[1]{\texttt{#1}}
\newtheorem{proposition}{Proposition}
\newtheorem{corollary}{Corollary}
\theoremstyle{definition}
\newtheorem{definition}{Definition}

\title{Right Reason, Wrong Advice: Evaluating Recourse \\ in LLM Explanations of Denied Requests}

\author{Arunkumar V \\ University College of Engineering, BIT Campus, Anna University \\ Tiruchirappalli, India \\ \texttt{arunkumarv1530@gmail.com}}

\begin{document}
\maketitle

\begin{abstract}
Language models increasingly write the explanations people receive when a rule system denies their request. If several rules deny the request independently, an explanation can name a true reason and still leave the person denied after they fix it. We call the missing property recourse validity: changing what the explanation cites, without changing what was asked for, can flip the decision. We compute it exactly, and a hitting-set argument predicts from the policy alone when a correct reason fails. In \oracleNGapPorts{} of \oracleNPorts{} rule systems ported from access control and US benefit eligibility, a single correct reason almost never gives recourse when several reasons apply. Across \realNModelsAll{} LLMs, checks of whether the cited reasons are correct, computed or asked of an LLM judge, are close to chance at spotting this failure, as the argument predicts. A strong judge asked whether acting on the explanation would work does spot it. Stronger and reasoning models do better but still fail on many denials with four or more independent reasons. Asking for every rule helps every open model with room to improve. Explanations of eligibility decisions should be built from the decision function and judged by whether acting on them works.
% CHECK: oracle-control: false if Cedar github-model's goal-free or goal-fixed share left denied is not 0
% CHECK: v2-stratum: false if fidelity predicts recourse on single-reason denials below AUROC 0.8, or on multi-reason ones more than 5 points from one half
% CHECK: v2-auroc-correctness: false if fidelity, causal recall or precision separates recourse-valid explanations at an AUROC more than 5 points from one half
% CHECK: v2-judge-correctness: false if a judge's correctness rating separates recourse-valid explanations with AUROC of 0.7 or more
% CHECK: v2-judge-spread: false if the best judge's actionable AUROC is under 0.9 or the weakest judge's reaches 0.8
% CHECK: v2-capability: false if the capability index no longer predicts recourse validity
% CHECK: v2-think: false if thinking does not raise every variant by more than 5 points to within 0.1 of the lowest frontier model, or lowers its precision by more than 5 points
% CHECK: v2-think-four: false if every thinking variant reaches 0.8 at four or more reasons
% CHECK: v2-frontier-four: false if no frontier model falls below 0.9 at four or more goal-fixed reasons
% CHECK: v2-list-all-every-open: false if list-all does not raise some open model below the ceiling at p under one percent
\end{abstract}

\section{Introduction}
\label{sec:intro}

Rule systems decide many requests people make. Cedar and OPA Gatekeeper admit or reject API calls and Kubernetes pods. Benefit agencies apply eligibility rules for food assistance, public housing and student aid. The decision is exact, but the explanation a person reads is more and more often generated by an LLM from the policy text and the request \citep{inan2023llama,rebedea-etal-2023-nemo}. Data-protection law asks that automated decisions come with meaningful information about the logic involved \citep{wachter2017right,goodman2017european}, and work on algorithmic recourse asks more: that an explanation tell the person what they could change to obtain a different outcome \citep{wachter-2018-counterfactual,ustun-2019-actionable,karimi-2022-survey}.

Figure~\ref{fig:example} shows a case from our benchmark. A family applying for public housing has net assets above the limit and also owns a home suitable for occupancy, and under HUD's rules each fact bars admission on its own. Gemma-4-12B, given the rules and the application, explains that the family's assets exceed the limit. The explanation is true, cites the right regulation and names a rule that fired. A family that acts on it, brings its assets under the limit and reapplies is denied again, this time because it owns a home.

Existing evaluations reward this explanation. Faithfulness tests ask whether an explanation reflects the process behind the decision \citep{jacovi-goldberg-2020-towards,atanasova-etal-2023-faithfulness,parcalabescu-frank-2024-measuring}, and it does. Simulatability asks whether a reader can predict the decision \citep{hase-bansal-2020-evaluating,chen2024models}. Causal recall against an exact oracle \citep{v-2026-content} credits the best coverage of any single reason, so this explanation scores 1. An LLM judge asked whether the reasons are correct has no reason to object. None of these asks whether acting on the explanation changes the outcome.

We define that property and compute it exactly from the decision function, given the features an extractor reads from each explanation. An explanation is recourse-valid if some change to the features it cites, holding fixed what the requester is asking for, flips the decision; equivalently, the cited features contain a contrastive explanation restricted to what the requester may change. One reason gives recourse if and only if it shares a feature with every other goal-fixed reason (Proposition~\ref{prop:hit}, the hitting-set duality of formal explanation restated with goals), so the failure is predictable from the policy alone. In \oracleNGapPorts{} of \oracleNPorts{} ported rule systems, no single correct reason gives recourse in \oracleNoSingleMin{} to \oracleNoSingleMax{} of the denials with two or more goal-fixed reasons (\S\ref{sec:oracle}).

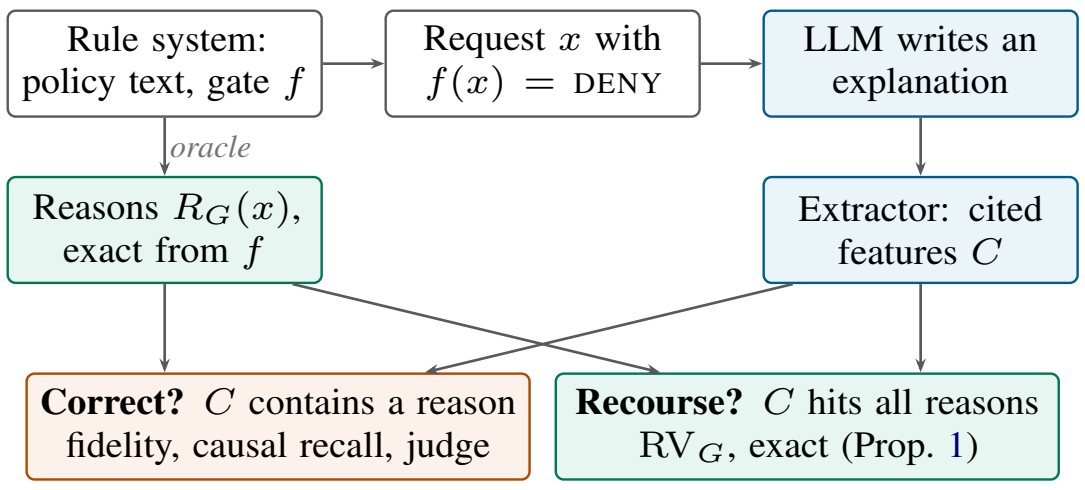
\begin{figure}[t]
\centering
% Measurement pipeline, column width. Colours follow fig_paper.py (Okabe-Ito).
\begin{tikzpicture}[
  font=\scriptsize, >={Stealth[length=2.6pt]},
  box/.style={draw=black!65, rounded corners=1.5pt, line width=0.45pt, align=center, inner sep=1.8pt,
              text width=#1, minimum height=7mm, fill=white},
  box/.default=19.5mm,
  arr/.style={->, line width=0.5pt, draw=black!65},
  lab/.style={font=\tiny\itshape, text=black!55, inner sep=0.8pt},
]
\node[box] (sys) at (0,0) {Rule system:\\ policy text, gate $f$};
\node[box] (ep) at (25mm,0) {Request $x$ with\\ $f(x)=\deny$};
\node[box, fill=oiBlue!8, draw=oiBlue!80!black] (llm) at (50mm,0) {LLM writes an\\ explanation};
\node[box, fill=oiGreen!9, draw=oiGreen!75!black] (rg) at (0,-11mm) {Reasons $R_G(x)$,\\ exact from $f$};
\node[box, fill=oiBlue!8, draw=oiBlue!80!black] (ext) at (50mm,-11mm) {Extractor: cited\\ features $C$};
\node[box=32mm, fill=oiVerm!8, draw=oiVerm!80!black] (corr) at (7.5mm,-24mm)
  {\textbf{Correct?} $C$ contains a reason\\ fidelity, causal recall, judge};
\node[box=32mm, fill=oiGreen!9, draw=oiGreen!75!black] (rec) at (42.5mm,-24mm)
  {\textbf{Recourse?} $C$ hits all reasons\\ $\RV_G$, exact (Prop.~\ref{prop:hit})};
\draw[arr] (sys) -- (ep);
\draw[arr] (ep) -- (llm);
\draw[arr] (llm) -- (ext);
\draw[arr] (sys) -- node[lab, right] {oracle} (rg);
\draw[arr] (rg.south) -- (rg.south |- corr.north);
\draw[arr] ($(rg.south east)+(-2mm,0)$) -- ($(rec.north west)+(7mm,0)$);
\draw[arr] (ext.south) -- (ext.south |- rec.north);
\draw[arr] ($(ext.south west)+(2mm,0)$) -- ($(corr.north east)+(-7mm,0)$);
\end{tikzpicture}
\caption{The oracle computes the goal-fixed reasons $R_G(x)$ from the gate, and an extractor reads the features $C$ an explanation cites. Correctness checks ask whether $C$ contains a reason; recourse validity asks whether $C$ hits all of them, which one reason does only if it shares a feature with every other (Proposition~\ref{prop:hit}).}
\label{fig:pipeline}
\end{figure}

We score \realNGensPaper{} explanations from \realNModelsAll{} LLMs, from 8B open models to frontier APIs (Figure~\ref{fig:pipeline}, Appendix~\ref{app:runs}). Model behaviour follows Corollary~\ref{cor:one}. Citing a true reason predicts recourse on single-reason denials (AUROC \realAurocSingleFidelity, small open models) and in the one system whose reasons share a feature (\realAurocFidelityGithub). On the other multi-reason denials every correctness signal sits near chance, with reasoning off or on, and recourse is predicted instead by how much of the reason structure an explanation covers (\S\ref{sec:correct}). LLM judges asked whether the reasons are correct pass nearly every explanation; a strong judge asked whether acting on it would work tracks the oracle (\S\ref{sec:judge}). Recourse validity tracks capability and reasoning: thinking lifts the \realNThink{} open models we ran with it from between \realThinkOffMin{} and \realThinkOffMax{} to between \realThinkOnMin{} and \realThinkOnMax{}. It does not remove the failure: at four or more independent reasons, \realFourNBelowEight{} of the \realFourNReasoning{} reasoning runs, thinking-on and frontier, fall below 0.8 (\S\ref{sec:who}). Asking for every rule narrows the gap for every open model that has one (\S\ref{sec:mitigations}), and advice works as written less often than recourse validity credits (\S\ref{sec:executed}).

Our contributions are:
\begin{enumerate}\setlength{\itemsep}{1pt}\setlength{\parskip}{0pt}
\item Recourse validity, a criterion that asks whether what a natural-language explanation cites contains a goal-restricted contrastive explanation, exact given those features (\S\ref{sec:construct}).
\item A benchmark of 420 episodes over \oracleNPorts{} ported rule systems, with goal and immutable features for each port (\S\ref{sec:benchmark}).
\item A measurement over \realNModelsAll{} LLMs. On multi-reason denials, correctness-style evaluation, including causal recall and an LLM judge asked about correctness, cannot tell explanations that give recourse from those that do not; a strong judge asked about the outcome can, and reasoning fails where independent reasons multiply (\S\ref{sec:results}).
\item Mitigations ranked by the gap they close, and a test of whether the advice works as written (\S\ref{sec:mitigations}, \S\ref{sec:executed}).
\end{enumerate}

\begin{figure}[t]
\centering
\small
\fbox{\parbox{0.94\columnwidth}{%
\textbf{Application} (HUD public housing). Income at or below 80\% of area median. Net family assets above the limit. Owns real property suitable for occupancy, no exception. No disqualifying history.\\[2pt]
\textbf{Rules.} F2: net assets over \$100,000 bar admission (24 CFR 5{.}618(a)(1)(i)). F3: owning suitable real property bars admission (5{.}618(a)(1)(ii)). Denied.\\[2pt]
\textbf{Explanation} (Gemma-4-12B, default prompt, verbatim). ``The application was denied because the net family assets exceed the permitted limit of \$100,000. This violation of eligibility requirements (24 CFR 5{.}618(a)(1)(i)) precludes admission regardless of other qualifying factors.''\\[2pt]
\textbf{Applicant.} Brings assets under the limit and reapplies. Denied again under F3.\\[2pt]
\textbf{Oracle.} The reasons $\{$\feat{net\_family\_assets}$\}$ and $\{$\feat{real\_property}$\}$ share no feature, so no single reason gives recourse. Fidelity 1, causal recall 1, $\RV=0$.}}
\caption{A correct explanation that gives no recourse, from the HUD port. Neither fact is immutable, so citing both would give actionable recourse.}
\label{fig:example}
\end{figure}

\section{Recourse validity}
\label{sec:construct}

\paragraph{Decisions and reasons.}
A rule system is a decision function $D:\mathcal{X}\to\{\allow,\deny\}$ over $\mathcal{X}=\prod_{f\in F}\mathcal{X}_f$, a product of finite feature domains. $D$ includes the policy's combination semantics. Under deny-overrides, as in Cedar, a request is allowed only if some permit matches and no forbid matches. Under violation-any, as in Gatekeeper, a request is rejected if any violation rule matches. Fix a context $x$ with $D(x)=\deny$. For $S\subseteq F$ let $[x]_S=\{y\in\mathcal{X}: y_S=x_S\}$. The set $S$ is sufficient at $x$ if $D(y)=\deny$ for every $y\in[x]_S$. The subset-minimal sufficient sets $\MSR(x)$ are the reasons for the denial, the abductive explanations of formal XAI \citep{shih2018symbolic,ignatiev2019abduction}. A rule-level oracle computes them from each ported policy.

\paragraph{Goals.}
A requester asks for something specific. A student applies for a Pell grant, a developer creates a pod, a user asks to edit a document. Let $G\subseteq F$ be the goal features that encode what is asked for. Changing a goal feature means asking for something else, so it is not recourse. For each port we name $G$ in Table~\ref{tab:oracle}.

\begin{definition}[Goal-fixed recourse validity, $\RV_G$]
Let an extractor map explanation $e$ to the features $C(e)\subseteq F$ it cites as reasons, and let $\Delta=C(e)\setminus G$ be the cited features the requester may change. Then
\[
\RV_G(e)=1 \Leftrightarrow \exists y\in[x]_{F\setminus\Delta}: D(y)=\allow
\]
This holds exactly when the kept features $F\setminus\Delta$ are not sufficient at~$x$, that is, when $\Delta$ contains a contrastive explanation of $x$ among the features the requester may change \citep{ignatiev2020contrastive}. $\RV$ is undefined if $C(e)$ is empty, and for the whole episode if $G$ alone is sufficient at~$x$, since no explanation can then give recourse.
\end{definition}

We write $\RV$ for $\RV_G$ where no confusion arises. $\RV$ is an existence condition. It credits an explanation if some assignment to the cited features flips the decision, even when the explanation does not say which, so it is an upper bound on how often acting on an explanation succeeds: when it fails, no change to what the explanation cites helps; when it succeeds, the explanation may still not lead the person to a working change. Section~\ref{sec:executed} measures the distance by executing the changes explanations propose. $\RV$ is also monotone in what is cited, since citing every changeable feature always succeeds, so we report causal precision beside it.

\paragraph{When one reason suffices.}
Define the goal-fixed reasons $R_G(x)$ as the subset-minimal members of $\{S\setminus G : S\in\MSR(x)\}$. A set $T\subseteq F\setminus G$ together with $G$ is sufficient iff $T$ contains a member of $R_G(x)$, and $R_G(x)=\{\emptyset\}$ exactly when $G$ alone forces the denial.

\begin{proposition}
\label{prop:hit}
Let $D(x)=\deny$ and let $G$ not be sufficient at $x$. Then $\RV(e)=1$ if and only if $\Delta$ meets every member of $R_G(x)$.
\end{proposition}

\begin{proof}[Proof sketch]
Let $K=F\setminus\Delta$, so $G\subseteq K$. If some $R\in R_G(x)$ misses $\Delta$, then $R=S\setminus G$ for some $S\in\MSR(x)$, so $S\subseteq R\cup G\subseteq K$, and $K$ is sufficient. Conversely, if $K$ is sufficient it contains some $S\in\MSR(x)$, and $S\setminus G$ contains some $R\in R_G(x)$ with $R\subseteq K\setminus G$, disjoint from $\Delta$.
\end{proof}

Proposition~\ref{prop:hit} is the goal-restricted form of the hitting-set duality between abductive and contrastive explanations \citep{ignatiev2020contrastive,marquessilva2022delivering}. We claim no new theory and use it only to predict when generated explanations fail.

\begin{corollary}
\label{cor:one}
If $e$ cites exactly one goal-fixed reason $R$, then $\RV(e)=1$ iff $R$ meets every other member of $R_G(x)$. If $R_G(x)$ has at least two members and they are pairwise disjoint, every single-reason explanation fails. If all members share a feature, a necessary feature in the sense of \citet{huang-2023-feature}, every single-reason explanation succeeds.
\end{corollary}

So the gap is visible from the policy before any model runs, and fixing the goal can create it: every privileged-containers violation requires that the request is not an update, so goal free one reason suffices (``make this an update''), and goal fixed none does (Table~\ref{tab:oracle}). % CHECK: oracle-goal-fixing: false if the privileged-containers share left denied is not 0 goal free and 1 goal fixed

\paragraph{Correctness measures.}
Fidelity asks whether an explanation discloses a reason the policy's trace reports (a matching forbid, a missing permit or a firing violation). The causal recall of \citet{v-2026-content} is the best coverage of any single reason, so one complete reason scores 1 and can still have $\RV=0$. Causal precision is the share of cited features that belong to some reason. Actionable recourse validity $\RV_A$ replaces $G$ with $G\cup I$, where $I$ are immutable features a requester cannot reasonably be told to change; it is undefined where $G\cup I$ alone forces the denial, so it covers fewer denials than $\RV_G$ and can exceed it.

\begin{table*}[t]
\centering
\footnotesize
\setlength{\tabcolsep}{3.4pt}
\begin{tabular}{lrrlrrcccrr}
\toprule
& & & & & Two+ & \multicolumn{3}{c}{Left denied} & No single & No \\
\cmidrule(lr){7-9}
Rule system & $|F|$ & Contexts & Goal $G$ & Denies & reasons & free & fixed & actionable & suffices & recourse \\
\midrule
Cedar document-cloud & 10 & 5,376 & \feat{action} & \oracleDocCloudDenies & \oracleDocCloudMultiGoal & \oracleDocCloudGapFree & \oracleDocCloudGapFixed & \oracleDocCloudGapActionable & \oracleDocCloudNoSingle & \oracleDocCloudNoRecourse \\
Cedar github-model & 7 & 704 & \feat{action} & \oracleGithubDenies & \oracleGithubMultiGoal & \oracleGithubGapFree & \oracleGithubGapFixed & \oracleGithubGapActionable & \oracleGithubNoSingle & \oracleGithubNoRecourse \\
Gatekeeper allowed repos & 5 & 576 & none & \oracleReposDenies & \oracleReposMultiGoal & \oracleReposGapFree & \oracleReposGapFixed & \oracleReposGapActionable & \oracleReposNoSingle & \oracleReposNoRecourse \\
Gatekeeper privileged & 9 & 1,152 & \feat{is\_update} & \oraclePrivilegedDenies & \oraclePrivilegedMultiGoal & \oraclePrivilegedGapFree & \oraclePrivilegedGapFixed & \oraclePrivilegedGapActionable & \oraclePrivilegedNoSingle & \oraclePrivilegedNoRecourse \\
SNAP eligibility & 10 & 3,456 & none & \oracleSnapDenies & \oracleSnapMultiGoal & \oracleSnapGapFree & \oracleSnapGapFixed & \oracleSnapGapActionable & \oracleSnapNoSingle & \oracleSnapNoRecourse \\
Title IV student aid & 10 & 3,456 & \feat{aid\_program} & \oracleTitleFourDenies & \oracleTitleFourMultiGoal & \oracleTitleFourGapFree & \oracleTitleFourGapFixed & \oracleTitleFourGapActionable & \oracleTitleFourNoSingle & \oracleTitleFourNoRecourse \\
HUD public housing & 10 & 2,304 & none & \oracleHudDenies & \oracleHudMultiGoal & \oracleHudGapFree & \oracleHudGapFixed & \oracleHudGapActionable & \oracleHudNoSingle & \oracleHudNoRecourse \\
\bottomrule
\end{tabular}
\caption{Rule systems and the oracle-level gap, exact over every context. Cedar and the eligibility rules use deny-overrides, Gatekeeper violation-any on 3-container pods. \emph{Two+ reasons}: denials with two or more goal-fixed reasons. \emph{Left denied}: share of (denial, reason) pairs in which citing that one correct reason leaves the request denied, over denials with two or more reasons (free) or two or more goal-fixed reasons (fixed; actionable also fixes the immutable features). \emph{No single suffices}: share of those denials in which no reason cited alone gives goal-fixed recourse. \emph{No recourse}: share that the goal and immutable features alone deny. Shares are under the uniform context measure, not population rates.}
\label{tab:oracle}
\end{table*}

\section{Benchmark}
\label{sec:benchmark}

\paragraph{Porting protocol.}
Each port is an executable gate over categorical features that abstract the source's predicates, keeps its source's combination semantics, and records every omitted provision with the reason in the port file. Cedar contributes the document-cloud and GitHub example policies \citep{cedar-examples}, Gatekeeper the allowed-repositories and privileged-containers templates \citep{gatekeeper-library}; a single-container pod can violate each template at most once, so pods have three container slots. The eligibility ports encode SNAP \citep[7 CFR 273;][]{usda-snap} for a one-person household, HUD public housing admission \citep[24 CFR 960.201, 960.204 and 5.603 to 5.618;][]{cfr-hud} and Title IV student aid \citep[34 CFR 668.32 to 668.35;][]{cfr-fsa}. A port script checks each provision against the outcome the regulation fixes and regenerates the port file byte for byte. Cedar's tinytodo policy and the single-container Gatekeeper ports admit no multi-reason denial as ported, and we exclude them. % CHECK: oracle-excluded: false if a port with no kept episodes has a multi-reason denial in oracle.json

\paragraph{Goals and episodes.}
The goal is \feat{action} for Cedar, \feat{aid\_program} for Title IV and \feat{is\_update} for the privileged-containers pod; the other ports have none. Immutable features follow the source text (Appendix~\ref{app:immutable}). Every port's context space is small enough to enumerate, and for each port we draw 30 multi-reason denials, which have two or more reasons with the goal free, and 30 single-reason controls uniformly without replacement from all its denials of each kind, with a fixed seed, 420 episodes in all. With the goal fixed, \realStratumMultiTwo{} of the 210 multi-reason episodes keep two or more reasons, \realStratumMultiOne{} keep one and \realStratumMultiNone{} have none, as do \realStratumSingleNone{} controls, so no explanation of those \realStratumMultiNone{} and \realStratumSingleNone{} can give recourse; Cedar github-model has only \oracleGithubMultiGoal{} contexts with two or more goal-fixed reasons, and \oracleGithubKeptMultiGoal{} of its 30 sampled episodes are among them. Rates are over episodes with a defined verdict.

\paragraph{The oracle-level gap.}
\label{sec:oracle}
In \oracleNGapPorts{} of \oracleNPorts{} rule systems, no single correct reason gives recourse in \oracleNoSingleMin{} to \oracleNoSingleMax{} of the denials with two or more goal-fixed reasons, and a correct reason cited alone fails in \oracleGapMin{} to \oracleGapMax{} of (denial, reason) pairs (Table~\ref{tab:oracle}).
Cedar github-model is the negative control: its reasons share a role-membership feature, so by Corollary~\ref{cor:one} every single reason succeeds, and the measured share is \oracleGithubGapFixed{}. % CHECK: oracle-control: false if Cedar github-model's goal-free or goal-fixed share left denied is not 0
In the eligibility rule sets each disqualifying condition denies on its own, so \oracleSnapMultiGoal{} of \oracleSnapDenies{} SNAP denials and \oracleTitleFourMultiGoal{} of \oracleTitleFourDenies{} Title IV denials have two or more reasons, and with immutable features fixed \oracleNoRecourseEligMin{} to \oracleNoRecourseEligMax{} of their multi-reason denials admit no actionable recourse at all.

\section{Models, conditions and extraction}
\label{sec:setup}

\paragraph{Conditions.}
Every prompt gives the request, the decision and the policy with one sentence stating its combination semantics. The default asks for 2 to 4 sentences for the audit log (Appendix~\ref{app:prompts}). \emph{List-all} also asks for every rule or missing permission that causes the denial. \emph{Requester} asks for an explanation to the person who made the request, as long as it needs to be; we run it on the \realNSmall{} small open models, Haiku and GPT-6 Luna, \realRequesterNModels{} in all. \emph{Free} and \emph{free-all} show the request and decision but not the policy, as when an LLM narrates a decision made by a system it cannot inspect. Section~\ref{sec:mitigations} adds four conditions on three small models: uncapped list-all, a counterfactual prompt that asks what the person should change, the oracle's minimum change, and every reason (Appendix~\ref{app:prompts}).

\paragraph{Models.}
We run \realNModelsAll{} models in three groups. \emph{Open, 8B to 21B} (\realNSmall{}): Llama-3.1-8B, Granite-4.2-8B, gpt-oss-20b, Mistral-Nemo-12B, Ministral-3-14B, Gemma-3-12B, Gemma-4-12B, Qwen2.5-14B, Qwen3-14B \citep{yang-2025-qwen3}, Qwen3.5-9B and Phi-4-14B. \emph{Open, 27B to 32B} (\realNLarge{}): Gemma-3-27B, Gemma-4-31B, Qwen3-32B and Qwen3.8-27B. \emph{Frontier} (\realNFrontier{}): Claude Haiku 4.5, Sonnet 5.5 and Opus 5.5, GPT-6 Luna and GPT-6.1 Sol. Every model answers each prompt 5 times. Open models run 4-bit through ollama on two T4 GPUs at temperature 0.8, with at most 400 new tokens and thinking off; \realCapHitMaxPct{} of \realCapHitMaxModel{}'s answers reach that cap and at most \realCapHitRestMaxPct{} of any other model's. gpt-oss-20b \citep{openai-2025-gptoss} cannot turn reasoning off, so it runs at low reasoning effort with a 1,500-token budget that includes the hidden reasoning; \realGptOssAtBudgetPct{} of its answers reach it, the longest at \realGptOssMaxTokens{} tokens. % CHECK: v2-gptoss-budget: false if some gpt-oss answer reaches its token budget
Frontier models run through vendor interfaces at low reasoning effort (Haiku at its default) without temperature control. Large and frontier models answer the 210 multi-reason denials, the others all 420 episodes. With 8-bit weights, default and list-all recourse validity of Llama-3.1-8B and Gemma-3-12B move by \realQEightDiffMin{} to \realQEightDiffMax{}, so 4-bit quantisation does not drive the small models' rates. % CHECK: v2-q8: false if 8-bit weights move a model's recourse validity by more than 0.1

\paragraph{Extraction.}
Qwen3-14B in JSON mode at temperature 0 lists the features an explanation presents as reasons. A second extractor, \realAgreeModel{}, re-extracted a sample of explanations from each of the \realAgreeNModelsCovered{} models. It agrees with the first at Krippendorff's $\alpha=\realAgreeAlpha$ (micro-F1 \realAgreeMicroFOne{}) over \realAgreeN{} explanations from six models and at $\alpha=\realAgreeNewAlpha$ over \realAgreeNewN{} from the other \realAgreeNewNModels{}, where recourse validity under its labels moves by at most \realAgreeNewRecourseDiffMax{} per model. % CHECK: v2-extractor-new: false if second-extractor alpha on the newer models is below two thirds or shifts recourse by more than 0.1
One author labelled \realHumanN{} explanations stratified over models, conditions and systems; agreement with the extractor is $\alpha=\realHumanXAlpha$ (micro-F1 \realHumanMicroFOne{}), and the two give the same recourse verdict on \realHumanVerdictAgreePct{} of items, $\kappa=\realHumanKappa$ \realHumanKappaCI{}. The extractor errs toward validity (\realHumanExtractorOnly{} items valid only under its labels, \realHumanHumanOnly{} only under the human's), and recourse validity on the human labels is within \realHumanRecourseDiffMax{} of the extractor's in every condition. % CHECK: v2-human: false if human agreement is below two thirds or recourse on human labels differs by more than 0.1; CHECK: v2-human-kappa: false if item-level agreement on the verdict falls below kappa 0.8

\paragraph{Statistics.}
Rates are over explanations that cite at least one feature. Cells average per-episode means, since an episode's 5 samples share a prompt, with 95\% intervals from resampling episodes; AUROCs resample episodes too, and a within-episode AUROC fixes the reason count. Paired effects use sign-flip tests ($10^4$ permutations, Holm-corrected); correlations carry intervals from resampling models.

\section{Results}
\label{sec:results}

\subsection{Correctness checks miss the failure}
\label{sec:correct}

\begin{figure}[t]
\centering
\includegraphics[width=\columnwidth]{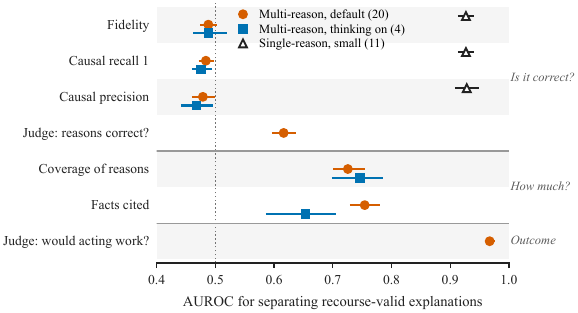}
\caption{How well each per-explanation signal separates recourse-valid explanations from the rest (AUROC, 95\% interval from resampling episodes; dotted line: chance). Circles: \realAurocN{} default explanations of multi-reason denials from all \realNModelsAll{} models; squares: \realThinkAurocN{} thinking-on explanations; triangles: single-reason controls, small open models. Judges are GPT-6 Luna on a sample (\S\ref{sec:judge}). Within-model and within-episode values are in Table~\ref{tab:signals}.}
\label{fig:signals}
\end{figure}

Table~\ref{tab:models} in Appendix~\ref{app:models} gives every model. Fidelity runs from \realCapFidMin{} to \realCapFidMax{} and recourse validity from \realMinDefault{} to \realFrontierMaxDefault{}.
Corollary~\ref{cor:one} says when a correct reason should predict recourse: on a single-reason denial, citing the true reason is what recourse needs; on a multi-reason denial whose reasons share no feature, it is not enough. Model behaviour follows the prediction. For the small open models, fidelity separates recourse-valid explanations at AUROC \realAurocSingleFidelity{} \realAurocSingleFidelityCI{} on the single-reason controls and at \realAurocMultiFidelity{} \realAurocMultiFidelityCI{} on the multi-reason denials. % CHECK: v2-stratum: false if fidelity predicts recourse on single-reason denials below AUROC 0.8, or on multi-reason ones more than 5 points from one half
Among the multi-reason systems it predicts recourse only on Cedar github-model, whose reasons share a feature, at \realAurocFidelityGithub{} \realAurocFidelityGithubCI{}, against \realAurocFidelityOtherMin{} to \realAurocFidelityOtherMax{} on the other six. % CHECK: v2-github-auroc: false if fidelity on Cedar github-model predicts recourse below AUROC 0.7, or another system reaches 0.7
Over all \realNModelsAll{} models, the three correctness signals separate recourse-valid explanations of multi-reason denials near chance (Figure~\ref{fig:signals}): fidelity at AUROC \realAurocFidelity{} \realAurocFidelityCI{}, causal recall at \realAurocRecall{} and causal precision at \realAurocPrecision{}. Within a model it ranges from \realAurocFidelityWithinMin{} to \realAurocFidelityWithinMax{}. % CHECK: v2-auroc-correctness: false if fidelity, causal recall or precision separates recourse-valid explanations at an AUROC more than 5 points from one half
The blind spot does not depend on how often models fail. On the \realThinkAurocN{} thinking-on explanations, \realThinkShareRV{} of them recourse-valid, fidelity separates at \realThinkAurocFidelity{} \realThinkAurocFidelityCI{} and causal recall at \realThinkAurocRecall{}, while coverage still reaches \realThinkAurocCoverage{}. % CHECK: v2-think-auroc: false if on thinking-on explanations fidelity or causal recall is more than 5 points from one half, or coverage is under 0.7
Of the \realRecallGapN{} explanations with causal recall 1, \realRecallGapPct{} leave the requester denied, up to \realRecallGapSmallMaxPct{} for \realRecallGapSmallMaxModel{}. % CHECK: v2-recall-gap: false if under a fifth of causal-recall-1 explanations leave the requester denied

Coverage of the reason structure predicts recourse instead. An explanation that covers one goal-fixed reason in full is recourse-valid in \realCoveredOneRV{} of cases, no more often than one that covers none (\realCoveredZeroRV); one that covers two or more is valid in \realCoveredTwoRV{}. Within an episode, coverage and the number of facts cited reach AUROC \realAurocCoverageWithinEp{} and \realAurocCitedWithinEp{}. In a logistic model with model fixed effects and episode-clustered errors, fidelity's coefficient is indistinguishable from zero once coverage and facts cited are known (\realLogitFidelityCoef, 95\% CI \realLogitFidelityCI), while both of those carry information (\realLogitCoverageP). Covering every reason implies recourse by Proposition~\ref{prop:hit}, so the coverage term is partly definitional; the informative result is the fidelity term. % CHECK: v2-covered-one: false if covering one reason differs from covering none by more than 5 points, or covering two or more is not 15 points above one; CHECK: v2-auroc-coverage: false if coverage or facts cited separates recourse-valid explanations at AUROC under 0.7; CHECK: v2-logit-fidelity: false if fidelity carries information beyond coverage and facts cited at p under five percent, or either of those does not

Two nulls separate which facts a model chooses from how many it cites. A random citation of as many facts of the episode gives less recourse than every model does, by \realAllMarginMin{} for \realAllMarginMinModel{} to \realAllMarginMax{} for \realAllMarginMaxModel{}; for \realCiTouchRandomModels{} the 95\% interval reaches the random rate. % CHECK: v2-random-all: false if some model's default recourse validity is at or below its size-matched random baseline
A random draw of as many facts from those that belong to some reason is the stronger null, and every one of the \realReasonNullNSmallBelow{} small open models falls below it, by as much as \realReasonNullSmallGapMax{} for \realReasonNullSmallMinModel{}; the frontier models sit at it (margins of \realReasonNullFrontierMin{} to \realReasonNullFrontierMax{}). % CHECK: v2-reason-null: false if some small open model reaches the reason-feature null, or a frontier model is more than 0.1 above it
The negative control behaves as predicted: on the \realControlTwoPlusN{} Cedar github-model episodes that keep two or more goal-fixed reasons, which share a feature, the small open models give recourse in \realControlTwoPlusSmallMean{} on average, and over all its episodes the system has the highest default recourse validity of the \oracleNPorts{} (Appendix~\ref{app:systems}). % CHECK: real-control: false if Cedar github-model's mean recourse validity under the policy condition is not the highest of the ported systems

\subsection{Who gives recourse}
\label{sec:who}

Recourse validity tracks capability. Across the \realCapEciN{} models with a published Epoch Capabilities Index \citep{epoch-2025-eci} it follows that index, Spearman \realCapEciRho{} \realCapEciCI{}, and across the \realCapEciOpenN{} open models alone \realCapEciOpenRho{} \realCapEciOpenCI{} (Appendix~\ref{app:eci}); among the \realCapEciSmallN{} small open models with an index it does not separate them (\realCapEciSmallRho, \realCapEciSmallP). % CHECK: v2-capability: false if the capability index no longer predicts recourse validity; CHECK: v2-eci-open: false if the open-model correlation is under 0.5 or p is at least five percent; CHECK: v2-eci-small: false if within the small open models the index predicts recourse at p under five percent
The frontier models give recourse in \realFrontierMinDefault{} to \realFrontierMaxDefault{} of multi-reason denials, the open models of 27B to 32B in \realLargeMinDefault{} to \realLargeMaxDefault{}, and those of 8B to 21B in \realMinDefault{} to \realMaxDefault{}. % CHECK: v2-small-below-frontier: false if a small open model reaches the lowest frontier model
Models of similar capability still differ widely. The \realCapBandN{} models with an index between \realCapBandLo{} and \realCapBandHi{} range from \realCapBandMin{} to \realCapBandMax{}: Qwen3-14B, Qwen3.5-9B and Qwen3-32B at the bottom, Haiku and Gemma-4-31B at the top. % CHECK: v2-band-spread: false if recourse validity spans less than 0.3 within the matched band; CHECK: v2-band-members: false if the band is no longer the three Qwen models at the bottom and Haiku and Gemma-4-31B at the top
Reasoning explains much of the band. With thinking on (gpt-oss-20b at high effort) and a 3,000-token budget that no answer reaches, Qwen3-14B, Qwen3-32B, Qwen3.8-27B and gpt-oss-20b rise from between \realThinkOffMin{} and \realThinkOffMax{} to between \realThinkOnMin{} and \realThinkOnMax{}, against \realFrontierMinDefault{} for the weakest frontier model. % CHECK: v2-think: false if thinking does not raise every variant by more than 5 points to within 0.1 of the lowest frontier model, or lowers its precision by more than 5 points; CHECK: v2-think-budget: false if some thinking-on answer reaches the 3,000-token budget
Length accounts for little of the gain: thinking adds \realThinkCitedGainMin{} to \realThinkCitedGainMax{} cited facts and moves causal precision by \realThinkPrecDiffMin{} to \realThinkPrecDiffMax{}.

\begin{figure}[t]
\centering
\includegraphics[width=\columnwidth]{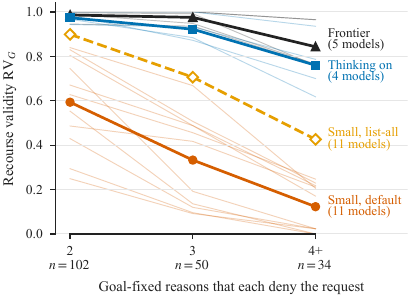}
\caption{$\RV_G$ by the number of goal-fixed reasons on the multi-reason denials ($n$ episodes per bucket). It falls in every group, reasoning models included. Bold: mean over a group's models; faint: one line per model. Thinking on: Qwen3-14B, Qwen3-32B, Qwen3.8-27B, and gpt-oss-20b at high effort. Per-bucket values are in Table~\ref{tab:reasons}.}
\label{fig:reasons}
\end{figure}

Recourse falls as reasons multiply, for every group (Figure~\ref{fig:reasons}). Under list-all, the small open models reach \realBucketTwoPolicyAll{} at two goal-fixed reasons and \realBucketFourPolicyAll{} at four or more, where a valid explanation must change \realBucketFourMinChange{} features on average. % CHECK: reasons-curve: false if ported list-all recourse does not fall from two to three to four or more goal-fixed reasons, or the synthetic gate's list-all rate is not above it at each of those counts
Reasoning does not remove this. At two goal-fixed reasons the thinking-on models give recourse in \realBucketTwoThinkMin{} to \realBucketTwoThinkMax{} of denials, at four or more in \realBucketFourThinkMin{} to \realBucketFourThinkMax{}; the frontier models average \realBucketFourFrontierPolicy{} there, the weakest \realBucketFourFrontierMin{}, and \realFourNBelowEight{} of the \realFourNReasoning{} reasoning runs fall below 0.8. % CHECK: v2-think-four: false if every thinking variant reaches 0.8 at four or more reasons, or one falls below 0.9 at two; CHECK: v2-frontier-four: false if no frontier model falls below 0.9 at four or more goal-fixed reasons
\citet{mayne-2025-decision} find no benefit from inference-time compute for self-generated counterfactuals; here reasoning helps until the independent reasons multiply, which is where Corollary~\ref{cor:one} says one reason, or two, stops being enough.
More reasoning effort narrows the fall for the strongest models only. Rerun at high effort on these denials, Sonnet gives recourse in \realHighFourSonnet{} \realHighFourSonnetCI{} and GPT-6 Luna in \realHighFourLuna{}, but Haiku in \realHighFourHaiku{}, up from \realBucketFourFrontierMin{} for all three at low effort, where GPT-6 Luna spent no reasoning tokens on a test prompt. % CHECK: v2-high-effort: false if high effort leaves every rerun model within 5 points of its low-effort value, lifts all three to 0.8 or more, or a rerun model did not start at the lowest frontier value

On the eligibility systems we also report $\RV_A$, which holds fixed what an applicant cannot change, such as citizenship. It is defined on only \realEligSnapNActionable{}, \realEligTitleFourNActionable{} and \realEligHudNActionable{} of the 30 sampled SNAP, Title IV and HUD denials, so these rates rest on few episodes: the small open models give actionable recourse in \realEligSmallTitleFourA{} of Title IV and \realEligSmallHudA{} of HUD denials on average, the open models of 27B to 32B in \realEligLargeTitleFourA{} and \realEligLargeHudA{}, and the frontier models in \realEligFrontierTitleFourA{} and \realEligFrontierHudA{}. % CHECK: v2-elig-actionable: false if the small open models' mean actionable recourse on Title IV or HUD reaches one half, or the frontier models' falls below 0.9

The strongest open model needs the policy text. Without it, Gemma-4-31B, the one model of 27B or more we ran without the policy, falls from \realGemmaNewBigPolicyRecourse{} to \realGemmaNewBigFreeRecourse{}, and on HUD from \realGemmaNewBigPolicyHudRecourse{} to \realGemmaNewBigFreeHudRecourse{}. % CHECK: v2-free-strong: false if Gemma-4-31B without the policy is within 0.2 of its policy-condition value

\subsection{Can an LLM judge stand in for the oracle?}
\label{sec:judge}

\begin{table}[t]
\centering
\small
\resizebox{\columnwidth}{!}{%
\begin{tabular}{lccccc}
\toprule
& \multicolumn{2}{c}{Correct} & Actionable & Quality & Difference \\
\cmidrule(lr){2-3}
Judge & 4+ & AUROC & AUROC & AUROC & AUROC \\
\midrule
GPT-6 Luna & \realJudgeLunaJudgeCorrectHighPct & \realJudgeLunaJudgeAurocCorrect & \realJudgeLunaJudgeAurocActionable & \realJudgeLunaJudgeAurocQuality & \realJudgeLunaJudgeActMinusQual \\
Gemma-4-31B & \realJudgeGemmaJudgeCorrectHighPct & \realJudgeGemmaJudgeAurocCorrect & \realJudgeGemmaJudgeAurocActionable & \realJudgeGemmaJudgeAurocQuality & \realJudgeGemmaJudgeActMinusQual \\
Qwen3.8-27B & \realJudgeQwenJudgeCorrectHighPct & \realJudgeQwenJudgeAurocCorrect & \realJudgeQwenJudgeAurocActionable & \realJudgeQwenJudgeAurocQuality & \realJudgeQwenJudgeActMinusQual \\
\bottomrule
\end{tabular}}
\caption{LLM judges on default explanations of multi-reason denials, 20 per model and system, \realJudgeLunaJudgeN{} per judge over \realJudgeLunaJudgeNModels{} models. \emph{4+}: share rated 4 or 5 on correct. AUROC: how well each 1 to 5 rating separates recourse-valid explanations from the rest; \emph{Difference}: actionable minus quality AUROC, paired.}
\label{tab:judge}
\end{table}

We gave \realJudgeNJudges{} LLM judges \citep{zheng-2023-judging} the case, the requester's fixed goal and one explanation, and asked in one prompt for 1 to 5 ratings of whether its reasons are true, whether changing exactly the facts it names would get the request allowed, and its overall quality (Appendix~\ref{app:prompts}). Asked whether the reasons are correct, the judges rate \realJudgeCorrectHighPctMin{} to \realJudgeCorrectHighPctMax{} of explanations 4 or 5, and the rating separates recourse-valid explanations at AUROC \realJudgeAurocCorrectMin{} to \realJudgeAurocCorrectMax{}, \realJudgeAurocCorrectWithinMin{} to \realJudgeAurocCorrectWithinMax{} within a model (Table~\ref{tab:judge}). % CHECK: v2-judge-correctness: false if a judge's correctness rating separates recourse-valid explanations with AUROC of 0.7 or more
It barely tracks fidelity either (AUROC \realJudgeCorrectFidMin{} to \realJudgeCorrectFidMax), so a judge asked about correctness passes nearly everything. % CHECK: v2-judge-correct-saturates: false if some judge rates under three quarters of explanations 4 or more on correct, or its correct rating separates faithful explanations at AUROC of 0.6 or more
Asked whether acting on the explanation would work, GPT-6 Luna reaches AUROC \realJudgeLunaJudgeAurocActionable{} \realJudgeLunaJudgeAurocActionableCI{} and Gemma-4-31B \realJudgeGemmaJudgeAurocActionable{} \realJudgeGemmaJudgeAurocActionableCI{}, and their mean ratings rank the models almost as the oracle does (Spearman \realJudgeLunaJudgeRhoActionable{} and \realJudgeGemmaJudgeRhoActionable); Qwen3.8-27B reaches \realJudgeQwenJudgeAurocActionable{}. % CHECK: v2-judge-spread: false if the best judge's actionable AUROC is under 0.9 or the weakest judge's reaches 0.8
In the joint prompt, quality asked right after the outcome question does about as well (Table~\ref{tab:judge}). % CHECK: v2-judge-quality: false if every judge's actionable AUROC beats quality by more than 5 points
Asked one question per call, on \realJudgeOnlyActionableN{} of the explanations it had already rated, Luna reaches \realJudgeOnlyActionableAlone{} \realJudgeOnlyActionableAloneCI{} on the outcome question, as in the joint prompt (\realJudgeOnlyActionableInThree), but \realJudgeOnlyQualityAlone{} \realJudgeOnlyQualityAloneCI{} on quality, down from \realJudgeOnlyQualityInThree{}: part of the quality rating's strength came from the question asked before it. % CHECK: v2-judge-only: false if the outcome question asked alone falls below 0.9, or quality asked alone is not at least 3 points below its joint-prompt value
Leaving out each judge's own explanations moves no AUROC by more than \realJudgeSelfShiftMax{}. % CHECK: v2-judge-self: false if leaving out a judge's own explanations moves some AUROC by more than 2 points
So the correctness question fails, and only the exact check shows which judge is strong enough to ask about the outcome.

\section{Mitigations}
\label{sec:mitigations}

\paragraph{Asking what to change does not help.}
On the \realNModelsAblate{} models of Table~\ref{tab:ladder} (Appendix~\ref{app:ladder}), a prompt that asks the model to tell the person exactly what would need to change leaves recourse validity within \realCounterfactualOverPolicyMin{} to \realCounterfactualOverPolicyMax{} of the default. % CHECK: real-counterfactual-no-help: false if some ladder model's counterfactual recourse validity differs from its default by 0.1 or more

\paragraph{Name every rule.}
One task sentence raises recourse validity for every open model whose default is below \realListAllCeiling{}, \realListAllNOpen{} models, by \realListAllDeltaMin{} for \realListAllDeltaMinModel{} to \realListAllDeltaMax{} for \realListAllDeltaMaxModel{} (paired sign-flip tests, every Holm-corrected $p<10^{-2}$), and the frontier models by at most \realListAllDeltaFrontierMax{}. % CHECK: v2-list-all-every-open: false if list-all does not raise some open model below the ceiling at p under one percent
That closes \realListAllGapShareMin{} of the default's gap for \realListAllGapShareMinModel{} to \realListAllGapShareMax{} for \realListAllGapShareMaxModel{}, under half for \realListAllNUnderHalf{} models. List-all raises every small open model, to between \realMinAll{} and \realMaxAll{}, and for each it gains more than its extra facts buy at random. % CHECK: real-list-all-raises: false if some model's pooled list-all recourse validity is at or below its policy-condition value; CHECK: real-gain-not-size: false if for some model the list-all gain in recourse validity is no larger than the gain of the size-matched random baseline
Reasoning and list-all add up. We ran list-all with thinking on for Qwen3-14B and gpt-oss-20b; it gives recourse in \realThinkAllMin{} to \realThinkAllMax{} of multi-reason denials, above either alone, and in \realThinkAllFourMin{} to \realThinkAllFourMax{} of those with four or more reasons, up from \realThinkAllBaseFourMin{} to \realThinkAllBaseFourMax{} for the same two models with thinking alone. % CHECK: v2-think-all: false if thinking list-all does not beat both thinking alone and thinking-off list-all for every variant, overall and at four or more reasons
The cost is length: the small open models' list-all explanations cite \realMeanPolicyAllCited{} facts on average against \realMeanPolicyCited{} by default, and causal precision falls for \realNPrecisionFall{} of the \realNModels{} small open models. % CHECK: real-cited-rises: false if list-all explanations cite no more facts on average than policy-condition ones

\paragraph{Address the person.}
The default asks for a short note for the audit log. Asked instead to explain the denial to the person who made the request, at any length, \realRequesterNRaise{} of the \realRequesterNSmall{} small open models we ran give more recourse, \realRequesterMeanGain{} more on average, and \realRequesterNBelowAll{} still give less than when asked for every rule, by \realRequesterMeanAllGap{} on average; \realRequesterFrontierModels{} move by at most \realRequesterFrontierDiffMax{}. % CHECK: v2-requester-mixed: false if addressing the person raises at most half or all of the small open models, or stays below list-all for at most half; CHECK: v2-requester-frontier: false if it moves a frontier model by more than 5 points
The audit-log framing explains less than half of the gap. % CHECK: v2-requester-part: false if the mean gain from addressing the person is half or more of the mean gap the default leaves
Most of the gain comes from length: on the three ladder models, addressing the person with the 2 to 4 sentence cap kept moves recourse validity by \realRequesterCapOverPolicyMin{} to \realRequesterCapOverPolicyMax{} from the default, and lifting the cap adds \realRequesterOverCapMin{} to \realRequesterOverCapMax{}. % CHECK: v2-requester-length: false if the capped requester prompt moves a ladder model by 5 points or more, or lifting the cap adds under 5 points

\paragraph{Give the model the reasons.}
A deployer that owns the decision function can put its reasons in the prompt. Only oracle content is both complete and precise: default and counterfactual prompts give less recourse than either oracle condition, and the list-all prompts less precision. % CHECK: real-oracle-complete-precise: false if for some ladder model the default or counterfactual recourse validity is not below both oracle conditions', or a list-all variant's precision is not below both oracle conditions'
Handed every reason, the three models average \realMeanOracleAllRecourse{}, the most of any rung for each and \realOracleAllOverMinhitMin{} to \realOracleAllOverMinhitMax{} above the minimum change, where the extractor misses a feature named verbatim in \realMinhitMissMinPct{} to \realMinhitMissMaxPct{} of explanations. % CHECK: real-oracle-top: false if for some ladder model oracle-all recourse validity is not above that of every other condition; CHECK: real-oracle-above-all: false if some model's pooled oracle-all recourse validity is at or below its list-all or free-all value

\section{Does the advice work as written?}
\label{sec:executed}

$\RV$ credits any change to the cited features. To test advice as written, a second extraction pass reads the concrete changes an explanation proposes, each a feature and a new value from its domain (Appendix~\ref{app:prompts}), and the decision function runs on the request with exactly those changes; a change to a goal feature is dropped. We run it on the three ladder models (Qwen3-14B, Gemma-3-12B, Llama-3.1-8B) under the counterfactual and minimum-change prompts of Section~\ref{sec:mitigations}, and on the requester prompt. Under the first, \realExecCounterfactualAdviceMin{} to \realExecCounterfactualAdviceMax{} of explanations propose a usable change, yet applying it gets the request allowed in only \realExecCounterfactualExecRVDefinedMin{} to \realExecCounterfactualExecRVDefinedMax{}, against recourse validity of \realExecCounterfactualExecRVMin{} to \realExecCounterfactualExecRVMax{} on the same explanations. % CHECK: v2-exec-optimistic: false if some counterfactual or minimum-change cell's executed share reaches four fifths of its recourse validity, or recourse-valid counterfactual advice works in four fifths or more
Advice from an explanation that fails $\RV$ works in at most \realExecCounterfactualExecGivenNotRVMax{} of cases, and from one that passes in \realExecCounterfactualExecGivenRVMin{} to \realExecCounterfactualExecGivenRVMax{}, so under this prompt $\RV$ is close to a necessary condition for working advice and an optimistic bound on it. % CHECK: v2-exec-necessary: false if counterfactual advice from explanations that fail RV works in 5 percent or more of cases for some model
Handed the minimum change, their advice works in \realExecMinhitExecRVDefinedMin{} to \realExecMinhitExecRVDefinedMax{} of explanations against $\RV$ of \realExecMinhitExecRVMin{} to \realExecMinhitExecRVMax{}, and advice from explanations that fail $\RV$ works in up to \realExecMinhitExecGivenNotRVMax{}. Every one of those \realExecMinhitOutsideN{} cases changes a fact the explanation offers as advice but does not present as a reason, so the reason extractor rightly leaves it out, since $\RV$ scores only the reasons an explanation gives. % CHECK: v2-exec-outside: false if some such advice changes only facts read as reasons
Addressed to the person, the \realRequesterNModels{} requester models give advice that works as written in \realExecRequesterExecutedMin{} to \realExecRequesterExecutedMax{} of explanations, against $\RV$ of \realExecRequesterExecRVMin{} to \realExecRequesterExecRVMax{}; for no model does it reach four fifths of what $\RV$ credits. % CHECK: v2-exec-requester: false if requester advice works as written in four fifths or more of its recourse validity for some model

\section{Related work}
\label{sec:related}

\paragraph{Faithfulness, sufficiency and judges.}
Faithfulness tests ask whether an explanation reflects what drove the explained model \citep{jacovi-goldberg-2020-towards,atanasova-etal-2023-faithfulness,parcalabescu-frank-2024-measuring}, and LLM explanations often leave out influential factors or do not suffice for the output \citep{turpin2023language,matton-2025-walk,arxiv-2606-28615}; \citet{pawar-2026-necessary} test whether changing one cited factor changes the LLM's own decision. Simulatability and utility studies ask whether a reader can predict outputs or solve new cases \citep{chen2024models,joshi-2023-machine}. Each measures an explanation against the model that produced it. We measure it against an external rule system, where independent reasons give every single factor zero necessity. LLM judges are lenient and near chance against objective labels \citep{zheng-2023-judging,tan-2025-judgebench,chen-2026-construct}, and they detect incomplete justifications of rule-governed decisions near chance \citep{zhao-2026-rgdt}; we find the same for the correctness question and show that the outcome question restores the signal. \citet{v-2026-content} score explanations of one synthetic tool-call gate against its exact minimal sufficient reasons; we use that gate only in Appendix~\ref{app:synthetic}, and its causal recall is one of the correctness signals we find uninformative about recourse.

\paragraph{Counterfactuals and recourse.}
Counterfactual explanations state what would have to change for a different outcome \citep{wachter-2018-counterfactual}, and recourse restricts the change to what a person can act on, including certifying when none exists \citep{ustun-2019-actionable,karimi-2022-survey,kothari-2024-prediction}; $\RV_G$ and $\RV_A$ carry these notions to the features a natural-language explanation cites. Reason codes assume a person can act on one reason at a time \citep{barocas-2020-hidden,cheon-2025-responsiveness}. LLMs' own counterfactuals are often invalid or far from minimal \citep{mayne-2025-decision,dehghanighobadi-etal-2025-llms}, less so in larger models \citep{kalyvas-2026-empirical}, but validity there is judged against the LLM's own boundary. RGDT-Bench scores justifications for completeness against a reference warrant \citep{zhao-2026-rgdt}. Recourse validity needs no reference and disagrees with completeness both ways: \realValidIncompletePct{} of the recourse-valid explanations we score are incomplete, and a complete explanation whose reasons rest on immutable facts gives no actionable recourse.

\paragraph{Rules, reasoning and formal explanations.}
Rule benchmarks score whether LLMs reach the right decision from written rules \citep{zhou-2025-rulearena,gui-2025-logicgame,servantez-2026-openexempt}; access-control systems have long computed what a denied requester could change \citep{kapadia-2004-know,chowdary-2026-extree}. We fix the decision with an exact gate and score whether an LLM's explanation of it gives recourse. Reasoning models collapse as puzzles and constraint problems grow \citep{shojaee-2025-illusion,lin-2025-zebralogic}; the fall at four or more reasons is the same shape for explanations. Proposition~\ref{prop:hit} restates the hitting-set duality of diagnosis and formal explanation \citep{reiter-1987-theory,ignatiev-2019-relating,ignatiev2020contrastive} with goal features fixed, and its shared-feature case is feature necessity \citep{huang-2023-feature}. Appendix~\ref{app:related} gives a longer account.

\section{Conclusion}
\label{sec:conclusion}

An explanation of a rule-based denial can cite a true reason and still fail the person who acts on it, and checks of whether the reasons are correct, computed or asked of an LLM judge, cannot see this. Across \realNModelsAll{} LLMs on \oracleNPorts{} ported rule systems, recourse validity tracks capability and reasoning, and falls as independent reasons multiply, for reasoning and frontier models too. Reasoning or asking for every rule closes much of the gap.
Explanations of eligibility decisions should be built from the decision function and evaluated by whether acting on them works.

\section*{Limitations}

\paragraph{Extraction and labels.} Every model-side number passes through an LLM extractor, and Qwen3-14B also extracts its own explanations. A second extractor agrees with it at $\alpha$ of \realAgreeNewAlpha{} to \realAgreeAlpha{} and one author's labels at \realHumanXAlpha{}; both move recourse validity by little, but one human annotator is weaker evidence than several, the extractors can share errors, and the pass that reads proposed changes in Section~\ref{sec:executed} has not been checked by hand.

\paragraph{An optimistic construct.} $\RV$ credits any assignment to the cited features, including ones the explanation never suggests, so it overstates how often acting on an explanation succeeds. Section~\ref{sec:executed} measures how far: recourse-valid counterfactual advice works as written in \realExecCounterfactualExecGivenRVMin{} to \realExecCounterfactualExecGivenRVMax{} of cases, so the gains it reports for a mitigation overstate gains in working advice. $\RV_A$ is defined on only \realEligHudNActionable{} to \realEligSnapNActionable{} of the 30 sampled denials per eligibility system, since the rest admit no actionable recourse, and whether models tell an applicant that no change can help is not measured.

\paragraph{Ports are abstractions.} Each port folds its source's predicates into a few categorical features. SNAP is modelled for a one-person household. The CFR lags statute, for SNAP after the 2025 reconciliation act and for Title IV after Workforce Pell, and the features abstract over the changed lists. Contexts are counted under the uniform measure, where most eligibility contexts are denials, so no count in Table~\ref{tab:oracle} is a population rate. Goal and immutable features are our choice, and a different choice changes which denials count as multi-reason or as having actionable recourse.

\paragraph{Models and decoding.} Open models run 4-bit with thinking off, while the public capability index scores provider-hosted weights; we test 8-bit weights on two models and thinking on four, under the default prompt only, so the other conditions describe models that do not reason before answering. Thinking also raises the token budget (400 new tokens off, 3,000 on), though no default answer of the thinking-off Qwen models comes near the cap (the longest is \realThinkOffMaxTokens{} tokens) and none of gpt-oss-20b's reaches its budget, so the budget does not explain the gain; three of the four thinking models are Qwen hybrids, % CHECK: v2-think-off-budget: false if a thinking-off Qwen answer reaches the 400-token cap
 and we did not read their reasoning traces, so we cannot say what reasoning changes. We dropped Qwen3.5-9B with thinking on because more than half of its answers ran out of the token budget before the explanation. The matched capability band separates model families as much as capability, and audience is separated from length on three small models only. Frontier models run at low reasoning effort through vendor interfaces without temperature control, and we reran three of them at high effort on the four-or-more-reason denials only; other settings may change their rates. Large and frontier models answer the multi-reason denials only. All prompts and rule systems are in English.

\paragraph{Judges and readers.} Three judges cannot map how judge quality varies across models, and judges saw a sample of 20 explanations per model and system. The simulated readers of Appendix~\ref{app:readers} are language models, not people.

\section*{Ethics statement}

\paragraph{Human labels and simulated readers.} One author labelled the extraction sample; no other person took part. The reader study uses language models as readers and collects no data from people.

\paragraph{Data and licences.} The Cedar examples and Gatekeeper library are Apache-2.0. The federal eligibility rules are public-domain works of the US government (17 U.S.C. 105). Every context is a synthetic feature combination. The eligibility ports are research abstractions, not eligibility determinations, and must not be used to advise applicants. Frontier models were called with provider data retention off where the interface allows it.

\paragraph{Release.} We will release the ports and port scripts, the episodes, every prompt, generation, extraction and judge rating, the oracle, and the analysis code that writes every number in this paper.

\paragraph{Dual use.} Disclosing every reason for an access-control denial tells a requester more about the policy, which can help an adversary probe it. For benefit denials the balance runs the other way. An applicant told one of three disqualifying reasons may fix it, reapply and be denied again, at a cost in time and lost benefits. Explanations of eligibility decisions should name every reason, or say that more than one applies.

\bibliography{references}

\appendix

\section{Extended related work}
\label{app:related}

\paragraph{Faithfulness.}
Faithfulness asks whether an explanation reflects how a decision was made \citep{jacovi-goldberg-2020-towards}. Tests insert reasons an explanation fails to mention or reconstruct the input from the stated reasons \citep{atanasova-etal-2023-faithfulness}, weight each mention by how far it moves the explained model \citep{siegel-2024-probabilities}, or check self-consistency, which is what most such tests measure \citep{parcalabescu-frank-2024-measuring,madsen-etal-2024-self}. Self-explanations are tuned for plausibility at some cost to faithfulness \citep{agarwal-2024-faithfulness}, and the concepts an explanation names can differ from those that move the output \citep{matton-2025-walk}. Chain-of-thought omits biasing features and hints the model used \citep{turpin2023language,chen-2025-reasoning}, models condition on their reasoning to different degrees \citep{lanham2023measuring,tutek-2025-measuring,gurarieh-2026-bonafide}, and LLM explanations are selective, naming a subset of causes as people do \citep{kunz-2024-properties}. Each of these compares an explanation with the model that produced it. In our setting the decision is made by an external rule system, so faithfulness to the explainer is not at issue: an explanation can be faithful to the rules and still give no recourse.

\paragraph{Sufficiency, simulatability and usefulness.}
Rationale benchmarks score comprehensiveness and sufficiency against the explained model \citep{deyoung-etal-2020-eraser}; with independent denial reasons, fixing one cited reason changes nothing, a failure these metrics cannot attribute to the explanation. Free-text LLM explanations are mostly insufficient for the model's own output \citep{arxiv-2606-28615}, and \citet{pawar-2026-necessary} test whether each cited factor is necessary or sufficient for an LLM's decision; under overdetermination every single factor has necessity zero, which Proposition~\ref{prop:hit} predicts from the policy. Simulatability asks whether an explanation helps a reader predict outputs \citep{hase-bansal-2020-evaluating,chen2024models,limpijankit-etal-2025-counterfactual}. Studies of human utility find that rationales help readers less than accuracy or similarity to gold rationales suggests \citep{joshi-2023-machine}, that few explanation datasets suit decision studies \citep{hashemichaleshtori-2024-evaluating}, and that readers over-rely on wrong explanations \citep{si-2024-large}. An incomplete but correct explanation carries the same risk, since the reader has no sign that a second reason exists. A position paper asks that self-explanations be judged on actionability \citep{arxiv-2607-15957}; recourse validity is one exact form of that judgement.

\paragraph{LLM judges.}
LLM judges agree with human raters on many generation tasks \citep{zheng-2023-judging,liu-2023-geval,chiang-2023-alternative}, but favour the response shown first \citep{wang-2024-fair}, their own outputs \citep{panickssery-2024-favor} and fluent answers with factual errors \citep{wu-2025-style}, and their reliability varies with the property rated and the wording of the question \citep{bavaresco-2025-instead,thakur-2025-judging,bellibatlu-2026-judgesense}. Against objective labels strong judges are close to chance \citep{tan-2025-judgebench}, and judges that ignore surface edits miss most edits that change what is being judged \citep{chen-2026-construct}. RGDT-Bench finds judges detect incomplete justifications of rule-governed decisions near chance \citep{zhao-2026-rgdt}. Our correctness question behaves the same way. We add two findings: asking whether acting on the explanation would work restores the signal for a strong judge, and the exact oracle shows which judge is strong enough.

\paragraph{Counterfactuals and recourse.}
People ask why an outcome happened rather than some other one \citep{miller-2019-explanation}. Counterfactual explanations state the smallest change that yields the preferred outcome \citep{wachter-2018-counterfactual}; methods make them feasible \citep{poyiadzi-2020-face} or diverse \citep{mothilal-2020-diverse}, and recourse restricts them to what a person can act on, holding immutable features fixed and certifying when no change helps \citep{ustun-2019-actionable,karimi-2022-survey,verma-2024-review,kothari-2024-prediction}. Recourse is owed to decision subjects as a matter of agency \citep{venkatasubramanian-2020-philosophical}. Reason codes rest on the assumption that features can be changed one at a time \citep{barocas-2020-hidden} and often name features whose change cannot alter the decision \citep{cheon-2025-responsiveness}, though in user studies with single-reason denials they help people obtain recourse as often as counterfactuals \citep{upadhyay-2025-recourse}. Our failure is the multi-reason case, where any single reason, coded or counterfactual, fails.

\paragraph{Law.}
The GDPR gives a right to be informed about the logic of automated decisions rather than an explanation of each \citep{wachter2017right}, and US explanation duties address inscrutable models more than scrutable ones \citep{selbst-2018-intuitive}; our rule systems are fully scrutable and their explanations still fail. Regulation~B requires adverse action notices to give the principal reasons and advises that more than four are unlikely to help the applicant \citep{cfpb-2026-regulationb}; a notice that respects that guidance can still give no recourse when more reasons deny independently.

\paragraph{LLMs and external decisions.}
LLMs' own counterfactuals are often invalid or far from minimal, judged against the LLM's own boundary \citep{mayne-2025-decision,dehghanighobadi-etal-2025-llms}, and their validity rises with scale \citep{kalyvas-2026-empirical}. LLMs also narrate the explanations of external classifiers \citep{zytek-2024-explingo} and propose recourse that a classifier checks inside the loop \citep{tabei-2026-agentic}. We score the free-text explanation a deployed model writes with no oracle in the loop. RGDT-Bench scores justifications of rule-governed decisions for completeness against reference warrants \citep{zhao-2026-rgdt}, and a SAT check can verify that a benefits justification is consistent with statute \citep{sunny-2026-neurosymbolic}.

\paragraph{Rules and policies.}
LLMs are tested on applying written rules: fees, trades and tax \citep{zhou-2025-rulearena}, rule-defined games \citep{gui-2025-logicgame}, statutes \citep{holzenberger-2020-dataset,blairstanek-2023-gpt3,guha-2023-legalbench} and bankruptcy exemptions generated from symbolic rules \citep{servantez-2026-openexempt}. Benefit agencies use LLMs to turn SNAP and Medicaid policy into code, which is accurate but incomplete \citep{kennan-2025-rules}; caseworkers follow chatbot answers on SNAP questions \citep{gosciak-2026-llms}; and LLMs draft Cedar policies under a verifier \citep{vatsa-2026-autocedar}. Guardrail systems explain refusals in natural language \citep{inan2023llama,rebedea-etal-2023-nemo}. Access-control systems have long computed the changes that would turn a denial into a grant \citep{kapadia-2004-know}, recently weighted by how changeable each attribute is \citep{chowdary-2026-extree}. These works score decisions or compute feedback; we score whether an LLM's explanation of a fixed decision gives recourse.

\paragraph{Reasoning.}
Open models expose reasoning as a switch, a token budget or an effort level \citep{yang-2025-qwen3,openai-2025-gptoss}. The accuracy of reasoning models falls as puzzles and constraint problems grow \citep{shojaee-2025-illusion,lin-2025-zebralogic}, longer reasoning can lower accuracy on deduction with many constraints \citep{gema-2025-inverse}, and chain-of-thought can hurt instruction following \citep{li-2025-thinking}. \citet{mayne-2025-decision} find no gain from inference-time compute for self-generated counterfactuals. Against an external rule system we find a large gain that fades as the number of independent reasons grows.

\paragraph{Formal explanations.}
Diagnoses are the minimal hitting sets of conflicts \citep{reiter-1987-theory}, and minimal unsatisfiable and minimal correction subsets are dual in the same way \citep{liffiton-2008-algorithms}. The duality was carried to classifier explanations \citep{ignatiev-2019-relating} and stated between abductive and contrastive explanations \citep{ignatiev2020contrastive,marquessilva2022delivering}. Related work defines sufficient and necessary reasons \citep{darwiche-2020-reasons}, selects among many abductive explanations \citep{audemard-2022-preferred}, and asks which features occur in every explanation \citep{huang-2023-feature}, the shared-feature case of Corollary~\ref{cor:one}. Actual causation defines explanations through causes \citep{halpern-2005-causes} and spreads responsibility when causes overdetermine an outcome \citep{chockler-2004-responsibility}; a denial with several disjoint goal-fixed reasons is that case, where no single cause is pivotal.

\section{Prompt wording}
\label{app:prompts}

Ported systems, semantics lines: ``A request is allowed only if some permit matches and no forbid matches.'' (Cedar, eligibility) and ``A request is rejected if any violation rule (FORBID) matches.'' (Gatekeeper). Default task line: ``Write 2-4 sentences for the audit log explaining why this request was denied.'' List-all adds: ``Name every rule or missing permission that causes the denial, not only one.'' Requester replaces the role sentence ``You write audit-log explanations for an access-control system.'' with ``You write explanations for the people whose requests an access-control system decides.'' and the task line with ``Write an explanation, for the person who made this request, of why it was denied. Make it as long as it needs to be.''; its capped variant uses ``Write 2-4 sentences, for the person who made this request, explaining why it was denied.'' Free and free-all drop the policy text and its semantics line; the free-all task line is ``Name every reason that causes the denial, not only one.'' Oracle-all keeps the policy prompt, appends ``Each of the following sets of facts is, on its own, enough for this denial:'' and one numbered line per minimal reason, such as ``(3) enrolled\_half\_time\_higher\_ed is yes and meets\_student\_exemption is no'' for a SNAP denial, and its task line is ``Explain every one of the reasons listed above, not only one.'' Uncapped list-all, counterfactual and minimum change keep the policy prompt. Uncapped list-all replaces the default task line with ``Write an explanation for the audit log, as long as it needs to be, of why this request was denied.'' Counterfactual replaces it with ``Write 2-4 sentences telling the person who made this request exactly what would need to change for it to be allowed.'' Minimum change appends ``Changing the following facts, with the request itself unchanged, would make this request allowed:'' followed by the oracle's minimum change as ``\textit{fact} (currently \textit{value})'', and its task line is ``Include the change listed above that would make it allowed.'' Eligibility prompts replace the access-control nouns with application nouns. Open-model generations are capped at 400 new tokens in every condition (gpt-oss-20b at 1,500 including reasoning).

The judge prompt gives the request context, the policy, its semantics line, the decision, ``The person cannot change what they are asking for:'' followed by the goal facts, and the explanation, then asks for three integers from 1 to 5: ``correct: Are the reasons the explanation gives true of this case under the policy?'', ``actionable: If the person changed exactly the facts the explanation names as reasons, and nothing else, would the request then be allowed?'' and ``quality: Overall, how good is this explanation for the person who was denied?'', in JSON at temperature 0.

The executed-recourse pass gives Qwen3-14B the request details, each detail's allowed values and the explanation, and asks it to ``List every change to the details that the explanation tells or suggests the person make, each as the detail's name and its new value, chosen from that detail's allowed values,'' including a change only if the explanation states or clearly implies the new value, in JSON at temperature 0.

\section{A synthetic first-match gate}
\label{app:synthetic}

GlassGate \citep{v-2026-content} is an agent tool-call gateway with 10 features and eight block rules ahead of a default allow under first-match. Of its 67,876 block contexts, 24,308 have two or more reasons, and one correct reason leaves the request blocked in \oracleSynGapFree{} of its multi-reason denials. We draw 100 multi-reason block episodes, 34 with two reasons, 33 with three and 33 with four. The fired rule is skewed toward R4, first in 45 episodes, and R7 and R8 never fire first. The \emph{first-match} condition uses the gate's own header; \emph{order-neutral} says every matching rule applies independently and the order carries no meaning; \emph{list-all} keeps that header and asks for every rule that applies. We show the block rules in identity and reversed order, plus one seeded shuffle for Haiku and Qwen3 under first-match. Ranges below are over the six models above the rule in Table~\ref{tab:synthetic}, or over the five open ones among them where the text says so; the four below the rule are reported in the table only.

\begin{table*}[t]
\centering
\small
\begin{tabular}{lcccccccc}
\toprule
& \multicolumn{4}{c}{First-match header} & \multicolumn{2}{c}{Order-neutral header} & \multicolumn{2}{c}{List-all} \\
\cmidrule(lr){2-5}\cmidrule(lr){6-7}\cmidrule(lr){8-9}
Model & Valid reason & Fidelity & Two or more & $\RV$ & Two or more & $\RV$ & Two or more & $\RV$ \\
\midrule
Llama-3.1-8B & \synLlamaEightFirstValid & \synLlamaEightFirstFidelity & \synLlamaEightFirstTwoPlus & \synLlamaEightFirstRecourse & \synLlamaEightNeutralTwoPlus & \synLlamaEightNeutralRecourse & \synLlamaEightAllTwoPlus & \synLlamaEightAllRecourse \\
Gemma-3-12B & \synGemmaTwelveFirstValid & \synGemmaTwelveFirstFidelity & \synGemmaTwelveFirstTwoPlus & \synGemmaTwelveFirstRecourse & \synGemmaTwelveNeutralTwoPlus & \synGemmaTwelveNeutralRecourse & \synGemmaTwelveAllTwoPlus & \synGemmaTwelveAllRecourse \\
Qwen3-14B & \synQwenThreeFirstValid & \synQwenThreeFirstFidelity & \synQwenThreeFirstTwoPlus & \synQwenThreeFirstRecourse & \synQwenThreeNeutralTwoPlus & \synQwenThreeNeutralRecourse & \synQwenThreeAllTwoPlus & \synQwenThreeAllRecourse \\
Phi-4-14B & \synPhiFourFirstValid & \synPhiFourFirstFidelity & \synPhiFourFirstTwoPlus & \synPhiFourFirstRecourse & \synPhiFourNeutralTwoPlus & \synPhiFourNeutralRecourse & \synPhiFourAllTwoPlus & \synPhiFourAllRecourse \\
Qwen2.5-14B & \synQwenTwoFiveFirstValid & \synQwenTwoFiveFirstFidelity & \synQwenTwoFiveFirstTwoPlus & \synQwenTwoFiveFirstRecourse & \synQwenTwoFiveNeutralTwoPlus & \synQwenTwoFiveNeutralRecourse & \synQwenTwoFiveAllTwoPlus & \synQwenTwoFiveAllRecourse \\
Claude Haiku 4.5 & \synHaikuFirstValid & \synHaikuFirstFidelity & \synHaikuFirstTwoPlus & \synHaikuFirstRecourse & \synHaikuNeutralTwoPlus & \synHaikuNeutralRecourse & \synHaikuAllTwoPlus & \synHaikuAllRecourse \\
\midrule
Gemma-3-4B & \synGemmaFourFirstValid & \synGemmaFourFirstFidelity & \synGemmaFourFirstTwoPlus & \synGemmaFourFirstRecourse & \synGemmaFourNeutralTwoPlus & \synGemmaFourNeutralRecourse & \synGemmaFourAllTwoPlus & \synGemmaFourAllRecourse \\
Llama-3.2-3B & \synLlamaThreeFirstValid & \synLlamaThreeFirstFidelity & \synLlamaThreeFirstTwoPlus & \synLlamaThreeFirstRecourse & \synLlamaThreeNeutralTwoPlus & \synLlamaThreeNeutralRecourse & \synLlamaThreeAllTwoPlus & \synLlamaThreeAllRecourse \\
Mistral-Nemo-12B & \synMistralFirstValid & \synMistralFirstFidelity & \synMistralFirstTwoPlus & \synMistralFirstRecourse & \synMistralNeutralTwoPlus & \synMistralNeutralRecourse & \synMistralAllTwoPlus & \synMistralAllRecourse \\
OLMo-2-13B & \synOlmoFirstValid & \synOlmoFirstFidelity & \synOlmoFirstTwoPlus & \synOlmoFirstRecourse & \synOlmoNeutralTwoPlus & \synOlmoNeutralRecourse & \synOlmoAllTwoPlus & \synOlmoAllRecourse \\
\bottomrule
\end{tabular}
\caption{Synthetic gate, 100 multi-reason block episodes, 1,000 explanations per cell (1,500 for Haiku and Qwen3 under first-match). \emph{Valid reason}: share that disclose at least one complete reason. \emph{Fidelity}: share that disclose the reason of the rule that fires first in the order shown. \emph{Two or more}: share that disclose at least two. $\RV$: recourse-valid share among explanations that cite at least one feature. Episode-level 95\% bootstrap intervals are at most \synTwoPlusCIWidthMax{} wide for two or more and \synRecourseCIWidthMax{} for $\RV$.}
\label{tab:synthetic}
\end{table*}

Under first-match, the six models above the rule disclose a complete valid reason in \synValidMinPct{} of explanations for \synValidMinModel{} and up to \synValidMaxPct{} for \synValidMaxModel{}, yet only \synRecourseMin{} of the explanations that cite anything are recourse-valid for \synRecourseMinModel{} and \synRecourseMax{} for \synRecourseMaxModel{}, so \synBlockedMinPct{} to \synBlockedMaxPct{} leave the requester blocked (Table~\ref{tab:synthetic}).
Size does not predict it: Gemma-3-4B reaches \synGemmaFourFirstRecourse{} against \synGemmaTwelveFirstRecourse{} for Gemma-3-12B, and Llama-3.2-3B matches Llama-3.1-8B. % CHECK: syn-size-not-monotone: false if Gemma-3-4B's first-match recourse validity is not above Gemma-3-12B's, or Llama-3.2-3B and Llama-3.1-8B differ by more than 5 points
Each of the six discloses a valid reason more often than it gives recourse. % CHECK: syn-right-not-actionable: false if some model's valid-reason share under first-match is at or below its recourse validity
The five open models among them disclose the reason of the rule that fires first in only \synFidelityOpenMinPct{} to \synFidelityOpenMaxPct{} of explanations and Haiku in \synHaikuFirstFidelityPct{}, while Haiku gives recourse in \synHaikuFirstRecourse{}. % CHECK: syn-haiku-first-match: false if Haiku's first-match fidelity is below 0.9 or its first-match recourse validity is not below its fidelity
The gap tracks how often a model discloses two or more reasons: \synTwoPlusMinModel{} does so in \synTwoPlusMinPct{} of explanations and has the lowest recourse validity of the six, \synTwoPlusMaxModel{} in \synTwoPlusMaxPct{} and the highest. % CHECK: syn-two-plus-tracks: false if the model with the fewest two-or-more disclosures does not have the lowest recourse validity, or the most not the highest
Rule order changes recourse validity although it never changes a decision: identity and reversed order give \synLlamaEightFirstRecourseOrderZero{} and \synLlamaEightFirstRecourseOrderOne{} for Llama-3.1-8B, \synQwenThreeFirstRecourseOrderZero{} and \synQwenThreeFirstRecourseOrderOne{} for Qwen3-14B, \synGemmaTwelveFirstRecourseOrderZero{} and \synGemmaTwelveFirstRecourseOrderOne{} for Gemma-3-12B and \synHaikuFirstRecourseOrderZero{} and \synHaikuFirstRecourseOrderOne{} for Haiku. % CHECK: syn-order: false if one of these four models changes by less than 5 points between the two orders

The order-neutral header and one instruction together close most of the gap for the five open models. % CHECK: syn-list-all-closes: false if list-all closes at most half the first-match gap for some open model
Asked to name every rule, they disclose two or more reasons in \synAllTwoPlusMinPct{} to \synAllTwoPlusMaxPct{} of explanations, and recourse validity rises to \synAllRecourseMin{} to \synAllRecourseMax{}; the instruction's own increment over the header alone is \synAllOverNeutralHighMin{} to \synAllOverNeutralHighMax{} for Qwen3-14B, Phi-4-14B and Qwen2.5-14B and \synAllOverNeutralLowMin{} to \synAllOverNeutralLowMax{} for Gemma-3-12B and Llama-3.1-8B.
The header alone helps four of the five: Qwen3-14B from \synQwenThreeFirstRecourse{} to \synQwenThreeNeutralRecourse{}, Phi-4-14B from \synPhiFourFirstRecourse{} to \synPhiFourNeutralRecourse{}, Qwen2.5-14B from \synQwenTwoFiveFirstRecourse{} to \synQwenTwoFiveNeutralRecourse{} and, least, Gemma-3-12B from \synGemmaTwelveFirstRecourse{} to \synGemmaTwelveNeutralRecourse{}; not Llama-3.1-8B, from \synLlamaEightFirstRecourse{} to \synLlamaEightNeutralRecourse{}. % CHECK: syn-neutral-helps: false unless the gap share closed by the order-neutral header is larger for Qwen3, Phi-4 and Qwen2.5 than for Gemma, positive for Gemma, and at most 0 for Llama
Its reasons share features, so list-all recourse on this gate stays at or above \synBucketAllMin{} from two reasons up, and at four or more reasons the smallest change has \synBucketFourMinChange{} features against \realBucketFourMinChange{} on the ported systems.

\section{Which reason models pick}
\label{app:which}

\begin{table}[t]
\centering
\footnotesize
\setlength{\tabcolsep}{3pt}
\resizebox{\columnwidth}{!}{%
\begin{tabular}{lcccc}
\toprule
& \multicolumn{2}{c}{First-match header} & \multicolumn{2}{c}{Order-neutral header} \\
\cmidrule(lr){2-3}\cmidrule(lr){4-5}
Model & First listed & Switch & Concentration & $p$ \\
\midrule
Llama-3.1-8B & \synLlamaEightFirstListed{} (\synLlamaEightFirstListedChance) & \synLlamaEightSwitch & \synLlamaEightNeutralConc{} (\synLlamaEightNeutralConcNull) & \synLlamaEightNeutralConcP \\
Gemma-3-12B & \synGemmaTwelveFirstListed{} (\synGemmaTwelveFirstListedChance) & \synGemmaTwelveSwitch & \synGemmaTwelveNeutralConc{} (\synGemmaTwelveNeutralConcNull) & \synGemmaTwelveNeutralConcP \\
Qwen3-14B & \synQwenThreeFirstListed{} (\synQwenThreeFirstListedChance) & \synQwenThreeSwitch & \synQwenThreeNeutralConc{} (\synQwenThreeNeutralConcNull) & \synQwenThreeNeutralConcP \\
Phi-4-14B & \synPhiFourFirstListed{} (\synPhiFourFirstListedChance) & \synPhiFourSwitch & \synPhiFourNeutralConc{} (\synPhiFourNeutralConcNull) & \synPhiFourNeutralConcP \\
Qwen2.5-14B & \synQwenTwoFiveFirstListed{} (\synQwenTwoFiveFirstListedChance) & \synQwenTwoFiveSwitch & \synQwenTwoFiveNeutralConc{} (\synQwenTwoFiveNeutralConcNull) & \synQwenTwoFiveNeutralConcP \\
Claude Haiku 4.5 & \synHaikuFirstListed{} (\synHaikuFirstListedChance) & \synHaikuSwitch & \synHaikuNeutralConc{} (\synHaikuNeutralConcNull) & \synHaikuNeutralConcP \\
\bottomrule
\end{tabular}}
\caption{Position and concentration on the synthetic gate. \emph{First listed}: among explanations that disclose exactly one valid rule, the share that disclose the rule listed first, chance $1/k$ in parentheses. \emph{Switch}: share of episodes whose most-disclosed rule changes when the order is reversed. \emph{Concentration}: one minus normalised entropy of disclosed reasons per cell, uniform null in parentheses, paired permutation $p$.}
\label{tab:which}
\end{table}

When several reasons are valid, which one does a model disclose? Under the first-match header, Haiku discloses the first listed valid rule in \synHaikuFirstListed{} of single-reason explanations against a chance rate of \synHaikuFirstListedChance{}, and its most-disclosed rule switches in \synHaikuSwitch{} of episodes when the order is reversed (Table~\ref{tab:which}); under that header the first listed valid rule is the deciding rule. The five open models follow position far less, from \synFirstListedOpenMin{} for \synFirstListedOpenMinModel{} to \synFirstListedOpenMax{} for \synFirstListedOpenMaxModel{}. % CHECK: syn-position: false if Haiku's first-listed share is within 30 points of chance or its switch rate is under half, or an open model's first-listed share reaches the midpoint between Haiku's share and its chance rate

Under the order-neutral header, \synNeutralHighCount{} of the five open models disclose two or more reasons in most explanations, \synNeutralHighMin{} to \synNeutralHighMax{}, but Gemma-3-12B and Llama-3.1-8B in only \synGemmaTwelveNeutralTwoPlusPct{} and \synLlamaEightNeutralTwoPlusPct{}. % CHECK: syn-two-groups: false unless exactly Qwen3, Phi-4 and Qwen2.5 disclose two or more in at least half the order-neutral explanations, Gemma and Llama stay under 35 percent there and reach at least half under list-all
Asked to list every rule, Gemma-3-12B and Llama-3.1-8B disclose two or more \synGemmaTwelveAllTwoPlusPct{} and \synLlamaEightAllTwoPlusPct{} of the time, so they can list the reasons but by default do not; they are also the two models the header helps least. % CHECK: syn-neutral-helps-least: false if the two open models with the fewest order-neutral two-or-more disclosures are not the two whose first-match gap the header closes least
Their disclosures concentrate on one reason well above the uniform null, as do those of Qwen3-14B and Qwen2.5-14B. % CHECK: syn-concentration: false if Gemma, Llama, Qwen3 or Qwen2.5 has order-neutral concentration at or below the null or p of at least 1 percent, or Phi-4 has p below 1 percent

The choice is shared across models. Taking each episode's most-disclosed single reason under the order-neutral header, Gemma-3-12B, Llama-3.1-8B and Qwen3-14B agree pairwise on \synAgreeGemmaTwelveLlamaEight{}, \synAgreeGemmaTwelveQwenThree{} and \synAgreeLlamaEightQwenThree{} of shared episodes, against \synAgreeGemmaTwelveLlamaEightChance{}, \synAgreeGemmaTwelveQwenThreeChance{} and \synAgreeLlamaEightQwenThreeChance{} under uniform choice; Haiku agrees with them near chance under first-match, \synAgreeFirstHaikuMin{} to \synAgreeFirstHaikuMax{}. % CHECK: syn-agreement: false if a Gemma, Llama or Qwen3 pair agrees less than 20 points above chance under the order-neutral header, or Haiku and one of them differ from chance by more than 15 points under first-match
The \synTieAgreeNPairs{} pairs of the five open models name the same single reason with probability \synTieAgreeMin{} to \synTieAgreeMax{} against \synTieAgreeChanceMin{} to \synTieAgreeChanceMax{} by chance. % CHECK: syn-agreement-tie-robust: false if a Gemma, Llama or Qwen3 pair's tie-free agreement under the order-neutral header is less than 20 points above chance

Models favour salient reasons over short ones. The five open models disclose a smallest reason in only \synSmallestMinPct{} of single-reason explanations for \synSmallestMinModel{} and \synSmallestMaxPct{} for \synSmallestMaxModel{}, below the \synSmallestUniformMinPct{} to \synSmallestUniformMaxPct{} expected under uniform choice, and the rule disclosed most is R8, writes to unknown domains, for every one of them, followed for Gemma-3-12B and Llama-3.1-8B by R7, bulk financial export by an intern; neither fires first in any episode. % CHECK: syn-salience: false if an open model picks a smallest reason at least as often as uniform choice would, R8 is not every open model's most disclosed rule, R7 is not second for Gemma and Llama, or R7 or R8 fires first in an episode
With R8 and R7 moved to the middle of the order-neutral list, single-reason explanations still name them most: Gemma-3-12B \synProbeGemmaTwelveREight{} and \synProbeGemmaTwelveRSeven{}, Llama-3.1-8B \synProbeLlamaEightREight{} and \synProbeLlamaEightRSeven{}, Qwen3-14B \synProbeQwenThreeREight{} and \synProbeQwenThreeRSeven{} (R8 then R7; $n=\synProbeGemmaTwelveN$, \synProbeLlamaEightN, \synProbeQwenThreeN). % CHECK: syn-probe-top-two: false if, with R8 and R7 listed mid-list, R8 and R7 are not the two most disclosed rules for some probed model

\section{Per-model results}
\label{app:models}

Table~\ref{tab:models} gives fidelity, causal precision, facts cited and recourse validity for every model on the 210 multi-reason denials, under the default prompt and list-all.

\begin{table*}[t]
\centering
\small
\resizebox{\textwidth}{!}{%
\begin{tabular}{lccccccccc}
\toprule
& \multicolumn{3}{c}{What is cited} & \multicolumn{2}{c}{Default, all systems} & \multicolumn{2}{c}{Default, eligibility} & List-all & Recall 1, \\
\cmidrule(lr){2-4}\cmidrule(lr){5-6}\cmidrule(lr){7-8}\cmidrule(lr){9-9}
Model & Fidelity & Precision & Facts & $\RV_G$ [95\% CI] & Random & $\RV_G$ & $\RV_A$ & $\RV_G$ & denied \\
\midrule
\multicolumn{10}{l}{\emph{Open, 8B to 21B}} \\
Llama-3.1-8B & \realLlamaEightPolicyFidelity & \realLlamaEightPolicyPrecision & \realLlamaEightPolicyCited & \realLlamaEightPolicyRecourse{} {\scriptsize\realLlamaEightPolicyRecourseCI} & \realLlamaEightPolicyRandom & \realLlamaEightPolicyEligRecourse & \realLlamaEightPolicyEligActionable & \realLlamaEightPolicyAllRecourse & \realRecallGapLlamaEightPct \\
Granite-4.2-8B & \realGranitePolicyFidelity & \realGranitePolicyPrecision & \realGranitePolicyCited & \realGranitePolicyRecourse{} {\scriptsize\realGranitePolicyRecourseCI} & \realGranitePolicyRandom & \realGranitePolicyEligRecourse & \realGranitePolicyEligActionable & \realGranitePolicyAllRecourse & \realRecallGapGranitePct \\
gpt-oss-20b & \realGptOssPolicyFidelity & \realGptOssPolicyPrecision & \realGptOssPolicyCited & \realGptOssPolicyRecourse{} {\scriptsize\realGptOssPolicyRecourseCI} & \realGptOssPolicyRandom & \realGptOssPolicyEligRecourse & \realGptOssPolicyEligActionable & \realGptOssPolicyAllRecourse & \realRecallGapGptOssPct \\
Mistral-Nemo-12B & \realMistralPolicyFidelity & \realMistralPolicyPrecision & \realMistralPolicyCited & \realMistralPolicyRecourse{} {\scriptsize\realMistralPolicyRecourseCI} & \realMistralPolicyRandom & \realMistralPolicyEligRecourse & \realMistralPolicyEligActionable & \realMistralPolicyAllRecourse & \realRecallGapMistralPct \\
Ministral-3-14B & \realMinistralPolicyFidelity & \realMinistralPolicyPrecision & \realMinistralPolicyCited & \realMinistralPolicyRecourse{} {\scriptsize\realMinistralPolicyRecourseCI} & \realMinistralPolicyRandom & \realMinistralPolicyEligRecourse & \realMinistralPolicyEligActionable & \realMinistralPolicyAllRecourse & \realRecallGapMinistralPct \\
Gemma-3-12B & \realGemmaTwelvePolicyFidelity & \realGemmaTwelvePolicyPrecision & \realGemmaTwelvePolicyCited & \realGemmaTwelvePolicyRecourse{} {\scriptsize\realGemmaTwelvePolicyRecourseCI} & \realGemmaTwelvePolicyRandom & \realGemmaTwelvePolicyEligRecourse & \realGemmaTwelvePolicyEligActionable & \realGemmaTwelvePolicyAllRecourse & \realRecallGapGemmaTwelvePct \\
Gemma-4-12B & \realGemmaNewPolicyFidelity & \realGemmaNewPolicyPrecision & \realGemmaNewPolicyCited & \realGemmaNewPolicyRecourse{} {\scriptsize\realGemmaNewPolicyRecourseCI} & \realGemmaNewPolicyRandom & \realGemmaNewPolicyEligRecourse & \realGemmaNewPolicyEligActionable & \realGemmaNewPolicyAllRecourse & \realRecallGapGemmaNewPct \\
Qwen2.5-14B & \realQwenTwoFivePolicyFidelity & \realQwenTwoFivePolicyPrecision & \realQwenTwoFivePolicyCited & \realQwenTwoFivePolicyRecourse{} {\scriptsize\realQwenTwoFivePolicyRecourseCI} & \realQwenTwoFivePolicyRandom & \realQwenTwoFivePolicyEligRecourse & \realQwenTwoFivePolicyEligActionable & \realQwenTwoFivePolicyAllRecourse & \realRecallGapQwenTwoFivePct \\
Qwen3-14B & \realQwenThreePolicyFidelity & \realQwenThreePolicyPrecision & \realQwenThreePolicyCited & \realQwenThreePolicyRecourse{} {\scriptsize\realQwenThreePolicyRecourseCI} & \realQwenThreePolicyRandom & \realQwenThreePolicyEligRecourse & \realQwenThreePolicyEligActionable & \realQwenThreePolicyAllRecourse & \realRecallGapQwenThreePct \\
Qwen3.5-9B & \realQwenThreeFivePolicyFidelity & \realQwenThreeFivePolicyPrecision & \realQwenThreeFivePolicyCited & \realQwenThreeFivePolicyRecourse{} {\scriptsize\realQwenThreeFivePolicyRecourseCI} & \realQwenThreeFivePolicyRandom & \realQwenThreeFivePolicyEligRecourse & \realQwenThreeFivePolicyEligActionable & \realQwenThreeFivePolicyAllRecourse & \realRecallGapQwenThreeFivePct \\
Phi-4-14B & \realPhiFourPolicyFidelity & \realPhiFourPolicyPrecision & \realPhiFourPolicyCited & \realPhiFourPolicyRecourse{} {\scriptsize\realPhiFourPolicyRecourseCI} & \realPhiFourPolicyRandom & \realPhiFourPolicyEligRecourse & \realPhiFourPolicyEligActionable & \realPhiFourPolicyAllRecourse & \realRecallGapPhiFourPct \\
\multicolumn{10}{l}{\emph{Open, 27B to 32B}} \\
Gemma-3-27B & \realGemmaTwentySevenPolicyFidelity & \realGemmaTwentySevenPolicyPrecision & \realGemmaTwentySevenPolicyCited & \realGemmaTwentySevenPolicyRecourse{} {\scriptsize\realGemmaTwentySevenPolicyRecourseCI} & \realGemmaTwentySevenPolicyRandom & \realGemmaTwentySevenPolicyEligRecourse & \realGemmaTwentySevenPolicyEligActionable & \realGemmaTwentySevenPolicyAllRecourse & \realRecallGapGemmaTwentySevenPct \\
Gemma-4-31B & \realGemmaNewBigPolicyFidelity & \realGemmaNewBigPolicyPrecision & \realGemmaNewBigPolicyCited & \realGemmaNewBigPolicyRecourse{} {\scriptsize\realGemmaNewBigPolicyRecourseCI} & \realGemmaNewBigPolicyRandom & \realGemmaNewBigPolicyEligRecourse & \realGemmaNewBigPolicyEligActionable & \realGemmaNewBigPolicyAllRecourse & \realRecallGapGemmaNewBigPct \\
Qwen3-32B & \realQwenThreeBigPolicyFidelity & \realQwenThreeBigPolicyPrecision & \realQwenThreeBigPolicyCited & \realQwenThreeBigPolicyRecourse{} {\scriptsize\realQwenThreeBigPolicyRecourseCI} & \realQwenThreeBigPolicyRandom & \realQwenThreeBigPolicyEligRecourse & \realQwenThreeBigPolicyEligActionable & \realQwenThreeBigPolicyAllRecourse & \realRecallGapQwenThreeBigPct \\
Qwen3.8-27B & \realQwenThreeEightPolicyFidelity & \realQwenThreeEightPolicyPrecision & \realQwenThreeEightPolicyCited & \realQwenThreeEightPolicyRecourse{} {\scriptsize\realQwenThreeEightPolicyRecourseCI} & \realQwenThreeEightPolicyRandom & \realQwenThreeEightPolicyEligRecourse & \realQwenThreeEightPolicyEligActionable & \realQwenThreeEightPolicyAllRecourse & \realRecallGapQwenThreeEightPct \\
\multicolumn{10}{l}{\emph{Frontier}} \\
Claude Haiku 4.5 & \realHaikuPolicyFidelity & \realHaikuPolicyPrecision & \realHaikuPolicyCited & \realHaikuPolicyRecourse{} {\scriptsize\realHaikuPolicyRecourseCI} & \realHaikuPolicyRandom & \realHaikuPolicyEligRecourse & \realHaikuPolicyEligActionable & \realHaikuPolicyAllRecourse & \realRecallGapHaikuPct \\
Claude Sonnet 5.5 & \realSonnetPolicyFidelity & \realSonnetPolicyPrecision & \realSonnetPolicyCited & \realSonnetPolicyRecourse{} {\scriptsize\realSonnetPolicyRecourseCI} & \realSonnetPolicyRandom & \realSonnetPolicyEligRecourse & \realSonnetPolicyEligActionable & \realSonnetPolicyAllRecourse & \realRecallGapSonnetPct \\
Claude Opus 5.5 & \realOpusPolicyFidelity & \realOpusPolicyPrecision & \realOpusPolicyCited & \realOpusPolicyRecourse{} {\scriptsize\realOpusPolicyRecourseCI} & \realOpusPolicyRandom & \realOpusPolicyEligRecourse & \realOpusPolicyEligActionable & \realOpusPolicyAllRecourse & \realRecallGapOpusPct \\
GPT-6 Luna & \realLunaPolicyFidelity & \realLunaPolicyPrecision & \realLunaPolicyCited & \realLunaPolicyRecourse{} {\scriptsize\realLunaPolicyRecourseCI} & \realLunaPolicyRandom & \realLunaPolicyEligRecourse & \realLunaPolicyEligActionable & \realLunaPolicyAllRecourse & \realRecallGapLunaPct \\
GPT-6.1 Sol & \realSolPolicyFidelity & \realSolPolicyPrecision & \realSolPolicyCited & \realSolPolicyRecourse{} {\scriptsize\realSolPolicyRecourseCI} & \realSolPolicyRandom & \realSolPolicyEligRecourse & \realSolPolicyEligActionable & \realSolPolicyAllRecourse & \realRecallGapSolPct \\
\bottomrule
\end{tabular}}
\caption{The 210 multi-reason denials, default prompt unless a column says otherwise. \emph{Fidelity}: discloses a reason the policy's trace reports. \emph{Facts}: mean facts cited. $\RV_G$: mean of per-episode means with its 95\% bootstrap interval. \emph{Random}: $\RV_G$ of as many random facts of the episode. \emph{Eligibility}: mean over SNAP, Title IV and HUD; $\RV_A$ also fixes immutable facts and is defined on fewer denials. \emph{Recall 1, denied}: share of explanations with causal recall 1 that leave the requester denied.}
\label{tab:models}
\end{table*}

\section{Per-system results}
\label{app:systems}

\begin{table}[t]
\centering
\small
\resizebox{\columnwidth}{!}{%
\begin{tabular}{lccccc}
\toprule
& \multicolumn{3}{c}{Default} & List-all & Control \\
\cmidrule(lr){2-4}\cmidrule(lr){5-5}\cmidrule(lr){6-6}
System & Fidelity & Two+ & $\RV_G$ & $\RV_G$ & $\RV_G$ \\
\midrule
Cedar doc-cloud & \realMeanPolicyDocCloudFidelity & \realMeanPolicyDocCloudTwoPlus & \realMeanPolicyDocCloudRecourse & \realMeanPolicyAllDocCloudRecourse & \realMeanPolicyDocCloudControlRecourse \\
Cedar github & \realMeanPolicyGithubFidelity & \realMeanPolicyGithubTwoPlus & \realMeanPolicyGithubRecourse & \realMeanPolicyAllGithubRecourse & \realMeanPolicyGithubControlRecourse \\
OPA repos pod & \realMeanPolicyReposFidelity & \realMeanPolicyReposTwoPlus & \realMeanPolicyReposRecourse & \realMeanPolicyAllReposRecourse & \realMeanPolicyReposControlRecourse \\
OPA privileged pod & \realMeanPolicyPrivilegedFidelity & \realMeanPolicyPrivilegedTwoPlus & \realMeanPolicyPrivilegedRecourse & \realMeanPolicyAllPrivilegedRecourse & \realMeanPolicyPrivilegedControlRecourse \\
SNAP & \realMeanPolicySnapFidelity & \realMeanPolicySnapTwoPlus & \realMeanPolicySnapRecourse & \realMeanPolicyAllSnapRecourse & \realMeanPolicySnapControlRecourse \\
Title IV & \realMeanPolicyTitleFourFidelity & \realMeanPolicyTitleFourTwoPlus & \realMeanPolicyTitleFourRecourse & \realMeanPolicyAllTitleFourRecourse & \realMeanPolicyTitleFourControlRecourse \\
HUD & \realMeanPolicyHudFidelity & \realMeanPolicyHudTwoPlus & \realMeanPolicyHudRecourse & \realMeanPolicyAllHudRecourse & \realMeanPolicyHudControlRecourse \\
\midrule
Pooled & \realMeanPolicyFidelity & \realMeanPolicyTwoPlus & \realMeanPolicyRecourse & \realMeanPolicyAllRecourse & \realMeanPolicyControlRecourse \\
\bottomrule
\end{tabular}}
\caption{Per system, mean over the \realNModels{} small open models. \emph{Two+}: share that disclose two or more reasons. \emph{Control}: the 30 single-reason denials per system.}
\label{tab:real}
\end{table}

Table~\ref{tab:real} gives the per-system means behind Section~\ref{sec:correct}. For the small open models the policy text cuts both ways: without it, default recourse changes by \realFreeOverPolicyMin{} for \realFreeOverPolicyMinModel{} to \realFreeOverPolicyMax{} for \realFreeOverPolicyMaxModel{}, while fidelity drops by \realFreeFidelityDropMin{} to \realFreeFidelityDropMax{}. % CHECK: v2-free-small-mixed: false if dropping the policy moves every small open model's default recourse the same way; CHECK: real-free-correctness-cost: false if some model's pooled fidelity without the policy is not below its fidelity with it
The supplementary file \texttt{supplementary\_real\_cells.csv} gives per-model fidelity, coverage, two or more disclosed reasons, goal-free, goal-fixed and actionable recourse validity, causal precision and facts cited for every system, condition and stratum, \realNCells{} cells, each as a mean of per-episode means with its 95\% bootstrap interval over episodes.

\section{Recourse by number of reasons}
\label{app:reasons}

The values behind Figure~\ref{fig:reasons} are in Table~\ref{tab:reasons}, with the mean size of the smallest change in each bucket.

\begin{table}[t]
\centering
\small
\resizebox{\columnwidth}{!}{%
\begin{tabular}{lccccc}
\toprule
& \multicolumn{2}{c}{Small open} & Thinking & Frontier & \\
\cmidrule(lr){2-3}\cmidrule(lr){4-4}\cmidrule(lr){5-5}
$|R_G(x)|$ & Default & List-all & Default & Default & Change \\
\midrule
1 & \realBucketOnePolicy & \realBucketOnePolicyAll & \realBucketOneThinkMean & \realBucketOneFrontierPolicy & \realBucketOneMinChange \\
2 & \realBucketTwoPolicy & \realBucketTwoPolicyAll & \realBucketTwoThinkMean & \realBucketTwoFrontierPolicy & \realBucketTwoMinChange \\
3 & \realBucketThreePolicy & \realBucketThreePolicyAll & \realBucketThreeThinkMean & \realBucketThreeFrontierPolicy & \realBucketThreeMinChange \\
4+ & \realBucketFourPolicy & \realBucketFourPolicyAll & \realBucketFourThinkMean & \realBucketFourFrontierPolicy & \realBucketFourMinChange \\
\bottomrule
\end{tabular}}
\caption{$\RV_G$ by number of goal-fixed reasons, the values behind Figure~\ref{fig:reasons}: mean over the \realNModels{} small open models on both strata (\realBucketOneN{}, \realBucketTwoN{}, \realBucketThreeN{} and \realBucketFourN{} episodes, the \realStratumDefined{} of 420 with a defined verdict) and over the \realNThink{} thinking-on and \realNFrontier{} frontier models on the multi-reason stratum. \emph{Change}: mean size of a minimum hitting set of $R_G(x)$.}
\label{tab:reasons}
\end{table}

\section{Signal AUROCs}
\label{app:signals}

Table~\ref{tab:signals} gives the AUROCs behind Figure~\ref{fig:signals}, pooled, within model and within episode.

\begin{table}[t]
\centering
\footnotesize
\setlength{\tabcolsep}{4pt}
\begin{tabular}{lcccc}
\toprule
& & & \multicolumn{2}{c}{Within} \\
\cmidrule(lr){4-5}
Signal & AUROC & 95\% CI & model & episode \\
\midrule
\multicolumn{5}{l}{\emph{Is what it cites correct?}} \\
Fidelity & \realAurocFidelity & {\scriptsize\realAurocFidelityCI} & \realAurocFidelityWithin & \realAurocFidelityWithinEp \\
Causal recall 1 & \realAurocRecall & {\scriptsize\realAurocRecallCI} & \realAurocRecallWithin & \realAurocRecallWithinEp \\
Causal precision & \realAurocPrecision & {\scriptsize\realAurocPrecisionCI} & \realAurocPrecisionWithin & \realAurocPrecisionWithinEp \\
\multicolumn{5}{l}{\emph{How much does it cover?}} \\
Two+ reasons & \realAurocTwoPlus & {\scriptsize\realAurocTwoPlusCI} & \realAurocTwoPlusWithin & \realAurocTwoPlusWithinEp \\
Coverage & \realAurocCoverage & {\scriptsize\realAurocCoverageCI} & \realAurocCoverageWithin & \realAurocCoverageWithinEp \\
Facts cited & \realAurocCited & {\scriptsize\realAurocCitedCI} & \realAurocCitedWithin & \realAurocCitedWithinEp \\
\bottomrule
\end{tabular}
\caption{Signal AUROCs behind Figure~\ref{fig:signals}: how well each per-explanation signal separates recourse-valid explanations from the rest, over \realAurocN{} default explanations of multi-reason denials from all \realNModelsAll{} models. The interval resamples episodes. \emph{Within model}: mean of per-model AUROCs, which a signal cannot raise by telling models apart. \emph{Within episode}: mean over the \realAurocWithinEpN{} episodes with both outcomes, which fixes the reason count. \emph{Two+ reasons}: cites at least two of the episode's reasons in full; \emph{Coverage}: share of the episode's reasons cited in full (both with the goal free).}
\label{tab:signals}
\end{table}

\section{Fidelity against recourse by model}
\label{app:fig}
\begin{figure}[t]
\centering
\includegraphics[width=\columnwidth]{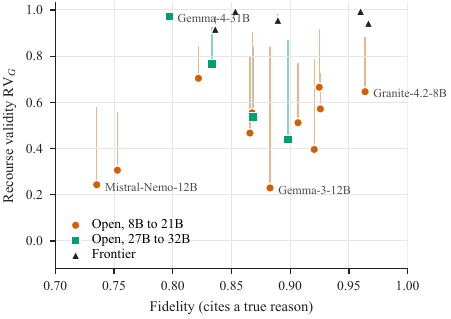}
\caption{Fidelity against goal-fixed recourse validity on the 210 multi-reason denials, one point per model, default prompt (Spearman \realCapFidRho{} over \realCapFidN{} models). Grey lines rise to each model's list-all value.}
\label{fig:models}
\end{figure}

Across models, fidelity and recourse validity are uncorrelated, Spearman \realCapFidRho{} \realCapFidCI{} (Figure~\ref{fig:models}). % CHECK: v2-fidelity-uninformative: false if fidelity and recourse validity correlate across models

\section{Prompt and content ladder}
\label{app:ladder}
Lifting the sentence limit adds \realPolicyAllNocapOverPolicyAllMin{} to \realPolicyAllNocapOverPolicyAllMax{} to list-all but lowers precision for every model. % CHECK: real-nocap-tradeoff: false if for some ladder model uncapped list-all has lower recourse validity than capped list-all, or precision that is not lower

\begin{table*}[t]
\centering
\small
\resizebox{\textwidth}{!}{%
\begin{tabular}{lccccccccc}
\toprule
& \multicolumn{6}{c}{Policy shown} & No policy & \multicolumn{2}{c}{Oracle content} \\
\cmidrule(lr){2-7}\cmidrule(lr){8-8}\cmidrule(lr){9-10}
Model & Default & Requester & Capped & List-all & Uncapped & Counterfactual & Free-all & Min.\ change & Reasons \\
\midrule
Qwen3-14B & \realQwenThreePolicyRecourse{} {\scriptsize\realQwenThreePolicyPrecision} & \realQwenThreeRequesterRecourse{} {\scriptsize\realQwenThreeRequesterPrecision} & \realQwenThreeRequesterCapRecourse{} {\scriptsize\realQwenThreeRequesterCapPrecision} & \realQwenThreePolicyAllRecourse{} {\scriptsize\realQwenThreePolicyAllPrecision} & \realQwenThreePolicyAllNocapRecourse{} {\scriptsize\realQwenThreePolicyAllNocapPrecision} & \realQwenThreeCounterfactualRecourse{} {\scriptsize\realQwenThreeCounterfactualPrecision} & \realQwenThreeFreeAllRecourse{} {\scriptsize\realQwenThreeFreeAllPrecision} & \realQwenThreeMinhitRecourse{} {\scriptsize\realQwenThreeMinhitPrecision} & \realQwenThreeOracleAllRecourse{} {\scriptsize\realQwenThreeOracleAllPrecision} \\
Gemma-3-12B & \realGemmaTwelvePolicyRecourse{} {\scriptsize\realGemmaTwelvePolicyPrecision} & \realGemmaTwelveRequesterRecourse{} {\scriptsize\realGemmaTwelveRequesterPrecision} & \realGemmaTwelveRequesterCapRecourse{} {\scriptsize\realGemmaTwelveRequesterCapPrecision} & \realGemmaTwelvePolicyAllRecourse{} {\scriptsize\realGemmaTwelvePolicyAllPrecision} & \realGemmaTwelvePolicyAllNocapRecourse{} {\scriptsize\realGemmaTwelvePolicyAllNocapPrecision} & \realGemmaTwelveCounterfactualRecourse{} {\scriptsize\realGemmaTwelveCounterfactualPrecision} & \realGemmaTwelveFreeAllRecourse{} {\scriptsize\realGemmaTwelveFreeAllPrecision} & \realGemmaTwelveMinhitRecourse{} {\scriptsize\realGemmaTwelveMinhitPrecision} & \realGemmaTwelveOracleAllRecourse{} {\scriptsize\realGemmaTwelveOracleAllPrecision} \\
Llama-3.1-8B & \realLlamaEightPolicyRecourse{} {\scriptsize\realLlamaEightPolicyPrecision} & \realLlamaEightRequesterRecourse{} {\scriptsize\realLlamaEightRequesterPrecision} & \realLlamaEightRequesterCapRecourse{} {\scriptsize\realLlamaEightRequesterCapPrecision} & \realLlamaEightPolicyAllRecourse{} {\scriptsize\realLlamaEightPolicyAllPrecision} & \realLlamaEightPolicyAllNocapRecourse{} {\scriptsize\realLlamaEightPolicyAllNocapPrecision} & \realLlamaEightCounterfactualRecourse{} {\scriptsize\realLlamaEightCounterfactualPrecision} & \realLlamaEightFreeAllRecourse{} {\scriptsize\realLlamaEightFreeAllPrecision} & \realLlamaEightMinhitRecourse{} {\scriptsize\realLlamaEightMinhitPrecision} & \realLlamaEightOracleAllRecourse{} {\scriptsize\realLlamaEightOracleAllPrecision} \\
\bottomrule
\end{tabular}}
\caption{Prompt and content ladder, $\RV_G$ on the 210 multi-reason denials with causal precision in small type. The oracle-content columns need a pipeline that computes the reasons from the decision function.}
\label{tab:ladder}
\end{table*}

\section{Runs}
\label{app:runs}

\begin{table}[t]
\centering
\footnotesize
\setlength{\tabcolsep}{4pt}
\begin{tabular}{@{}>{\raggedright\arraybackslash}p{2.75cm}r>{\raggedright\arraybackslash}p{3.2cm}@{}}
\toprule
Conditions & Episodes & Models \\
\midrule
\multicolumn{3}{@{}l}{\emph{Open, 8B to 21B}} \\
default, list-all & 420 & all 11 \\
requester & 210 & all 11 \\
requester, 2 to 4 sentences & 210 & Qwen3-14B, Gemma-3-12B, Llama-3.1-8B \\
free, free-all & 420 & 10, not gpt-oss-20b \\
uncapped, counterfactual, min.\ change & 210 & Qwen3-14B, Gemma-3-12B, Llama-3.1-8B \\
every reason & 420 & the same three \\
\addlinespace
\multicolumn{3}{@{}l}{\emph{Open, 27B to 32B}} \\
default, list-all & 210 & all 4 \\
free, free-all & 210 & Gemma-4-31B \\
\addlinespace
\multicolumn{3}{@{}l}{\emph{Frontier}} \\
default, list-all & 210 & all 5 \\
requester & 210 & Haiku, GPT-6 Luna \\
\addlinespace
\multicolumn{3}{@{}l}{\emph{Settings}} \\
default, list-all, 8-bit & 210 & Llama-3.1-8B, Gemma-3-12B \\
default, thinking on & 210 & Qwen3-14B, Qwen3-32B, Qwen3.8-27B, gpt-oss-20b \\
list-all, thinking on & 210 & Qwen3-14B, gpt-oss-20b \\
default, high effort & \realBucketFourN & Haiku, Sonnet, GPT-6 Luna \\
\bottomrule
\end{tabular}
\caption{Every generation run, 5 samples per episode and condition. The high-effort reruns cover only the \realBucketFourN{} multi-reason denials with four or more goal-fixed reasons. The \realNModelsAll{} models produced \realNGensPaper{} explanations: \realNGensSmall{} from the small open models, \realNGensLarge{} from the large and \realNGensFrontier{} from the frontier models; the settings runs are counted apart.}
\label{tab:runs}
\end{table}

Table~\ref{tab:runs} lists every run. Explanations without a parseable extraction are dropped and counted, \realDropped{} of \realNGens{}. The 210 episodes are the multi-reason stratum; the 420 add the single-reason controls.

\section{Capability index}
\label{app:eci}

\begin{table}[t]
\centering
\small
\begin{tabular}{lcc}
\toprule
Model & ECI & $\RV_G$ \\
\midrule
Llama-3.1-8B & \realEciLlamaEight & \realLlamaEightPolicyRecourse \\
Mistral-Nemo-12B & \realEciMistral & \realMistralPolicyRecourse \\
Gemma-3-12B & \realEciGemmaTwelve & \realGemmaTwelvePolicyRecourse \\
Gemma-3-27B & \realEciGemmaTwentySeven & \realGemmaTwentySevenPolicyRecourse \\
Phi-4-14B & \realEciPhiFour & \realPhiFourPolicyRecourse \\
gpt-oss-20b & \realEciGptOss & \realGptOssPolicyRecourse \\
Qwen3-14B & \realEciQwenThree & \realQwenThreePolicyRecourse \\
Qwen3-32B & \realEciQwenThreeBig & \realQwenThreeBigPolicyRecourse \\
Qwen3.5-9B & \realEciQwenThreeFive & \realQwenThreeFivePolicyRecourse \\
Claude Haiku 4.5 & \realEciHaiku & \realHaikuPolicyRecourse \\
Gemma-4-31B & \realEciGemmaNewBig & \realGemmaNewBigPolicyRecourse \\
GPT-6 Luna & \realEciLuna & \realLunaPolicyRecourse \\
Claude Sonnet 5.5 & \realEciSonnet & \realSonnetPolicyRecourse \\
GPT-6.1 Sol & \realEciSol & \realSolPolicyRecourse \\
Claude Opus 5.5 & \realEciOpus & \realOpusPolicyRecourse \\
\bottomrule
\end{tabular}
\caption{Epoch Capabilities Index values \citep{epoch-2025-eci}, read on 6 October 2026, and default goal-fixed recourse validity. Models without a published value are left out of the correlations.}
\label{tab:eci}
\end{table}

Table~\ref{tab:eci} lists the index values behind Section~\ref{sec:who}. The index scores provider-hosted weights and settings; our open models run 4-bit with thinking off.

\section{Simulated readers}
\label{app:readers}

LLM readers got stimuli written for a human study, and the oracle scored their answers. Each reader model saw every item under every condition five times at temperature 0.8, with the facts in a fresh random order, and chose the facts it would change to get the decision reversed while keeping its goal, or ``I cannot tell''. The 120 items are plain-language denials with two or more pairwise feature-disjoint goal-fixed reasons, 30 each from SNAP, HUD, Title IV and Cedar document-cloud. The \emph{one} explanation names one correct reason, \emph{all} every reason, and \emph{none} only that the request was denied. The study credits a choice of facts whatever values a reader would set, so it does not show that a reader finds a working change, and it is not evidence about how people read or act on explanations.
The reader models are \simReaders{}. Under \emph{one}, \simOneConfined{} of answers change only facts of the named reason. The chosen facts meet every reason in \simOneUnblockMin{} to \simOneUnblockMax{} of trials under \emph{one} and \simAllUnblockMin{} to \simAllUnblockMax{} under \emph{all}, so every reader does better with every reason than any does with one, % CHECK: sim-one-below-all: false if some reader does at least as well under one as some reader does under all
and \emph{all} differs from \emph{one} by \simHOneDiffMin{} to \simHOneDiffMax{} per reader, significant after Holm correction for \simHOneNSig{}.

\section{Immutable features}
\label{app:immutable}

Each port file lists the features that actionable recourse holds fixed with the goal, chosen from the source text. Document-cloud: \oracleDocCloudImmutable{}, as the policy compares the principal with the document's owner and no action changes the owner. GitHub: \oracleGithubImmutable{}, as who filed an issue is fixed when it is filed. SNAP: \oracleSnapImmutable{}, age 60 or disability and citizenship or eligible alien status \citep[7 CFR 273.9(a) and 273.4(a);][]{usda-snap}. HUD: \oracleHudImmutable{}, standards that ``permanently prohibit admission'', ``a lifetime registration requirement'' and eligible immigration status \citep[24 CFR 960.204(a)(3), (a)(4) and 5.506;][]{cfr-hud}. Title IV: \oracleTitleFourImmutable{}, no baccalaureate degree for a Pell Grant and citizenship or eligible noncitizen status \citep[34 CFR 690.75, 668.32(c) and 668.33(a);][]{cfr-fsa}. HUD's three-year drug-eviction bar lapses and has a rehabilitation exception, and Title IV's incarceration feature includes enrolling in an eligible prison education program, so both stay changeable. One port deviation is known: the document-cloud port applies the private-document forbid to document creation, where the source's owner guard means it never fires; it affects one control episode and no multi-reason episode.

\end{document}